\documentclass[lettersize,journal]{IEEEtran}
\usepackage{amsmath,amsfonts}
\makeatletter
\newif\if@restonecol
\makeatother

\usepackage[linesnumbered,ruled,vlined]{algorithm2e}
\usepackage{algpseudocode}
\usepackage{amsmath}
\usepackage{array}
\usepackage[caption=false,font=normalsize,labelfont=sf,textfont=sf]{subfig}
\usepackage{textcomp}
\usepackage{stfloats}
\usepackage{url}
\usepackage{verbatim}
\usepackage{graphicx}

\usepackage{threeparttable} 
\usepackage{booktabs}
\usepackage{multicol}
\usepackage{makecell} 
\renewcommand{\arraystretch}{1.2} 
\usepackage{multirow}
\usepackage{tabularx}

\usepackage{cite}
\usepackage{bm}
\usepackage{xfrac}
\usepackage{hyperref}
\hypersetup{
colorlinks=true, 
linkcolor=blue,  
citecolor=blue,  
urlcolor=blue,    
pdfborder={0 0 0} 
}
\begin{document}

\title{PIEDet: Prototype-Driven Intrinsically Explainable Object Detection}

\author{
	Jianlin~Xiang,
	Linhui~Dai$^{*}$,
	Xue~Yang,
	Chaolei~Yang,
	and~Yanshan~Li
	\thanks{This work was partially supported by National Natural Science Foundation of China (No.62471317), Natural Science Foundation of Shenzhen (No. JCYJ20240813141331042), and Guangdong Provincial Key Laboratory (Grant 2023B1212060076). \textit{($^{*}$Corresponding author: Linhui Dai.)}}
	\thanks{Jianlin Xiang, Linhui Dai, Chaolei Yang, and Yanshan Li are with the Institute of Intelligent Information Processing, Shenzhen University; the Guangdong Provincial Key Laboratory of Intelligent Information Processing, Shenzhen University; and the Shenzhen Key Laboratory of Modern Communications and Information Processing, Shenzhen University, Shenzhen, China 
		(e-mail: xiangjianlin2023@email.szu.edu.cn; dailinhui@szu.edu.cn; yangchaolei2022@email.szu.edu.cn; lys@szu.edu.cn).}
	\thanks{Xue Yang is with Shanghai Jiao Tong University, Shanghai, China (e-mail: yangxue-2019-sjtu@sjtu.edu.cn).}
}

\markboth{Journal of \LaTeX\ Class Files,~Vol.~14, No.~8, August~2021}%
{Shell \MakeLowercase{\textit{et al.}}: A Sample Article Using IEEEtran.cls for IEEE Journals}

\IEEEpubid{0000--0000/00\$00.00~````\copyright~2021 IEEE}

\maketitle

\begin{abstract}
	Existing object detectors typically make predictions in a black-box manner and struggle to simultaneously provide discriminative evidence for their predictions, which limits their deployment in safety-critical scenarios.
	To explain model predictions, existing post-hoc explanation methods mostly rely on gradient-based or perturbation-based operators. These methods not only introduce additional memory and computational overhead but also make it difficult to ensure that the generated explanations faithfully reflect the model’s internal decision-making process.
	To address these limitations, we propose PIEDet, a prototype-driven intrinsically explainable object detection framework. PIEDet innovatively embeds class prototypes as explicit discriminative units into the classification branch of a one-stage detector, thereby improving detection performance while providing intrinsic interpretability.
	First, PIEDet constructs hierarchical class prototypes at different detection levels, enabling the model to learn scale-aware class-semantic representations.
	Second, we propose a prototype-driven feature learning method consisting of prototype regularization and a region-to-prototype matching loss. The former enhances the inter-class discriminability of the prototypes, while the latter encourages prototype responses to focus on object regions.
	Finally, we introduce a scale-aligned hierarchical prototype supervision mechanism that assigns scale-matched supervision signals to different detection levels, thereby enhancing the scale specificity of the hierarchical prototypes.
	On the ExDark, RTTS, and VOC2012-FOG datasets, PIEDet improves mAP@0.5 over the baseline by 4.7\%, 1.6\%, and 4.8\%, respectively, while demonstrating superior computational efficiency. Compared with mainstream post-hoc explanation methods, PIEDet achieves a better balance between explanation quality and explanation cost.
	The code is available at https://github.com/xjlDestiny/PIEDet.git.
\end{abstract}


\section{Introduction}

\begin{figure}[!t]
	\centering
	\includegraphics[width=\columnwidth]{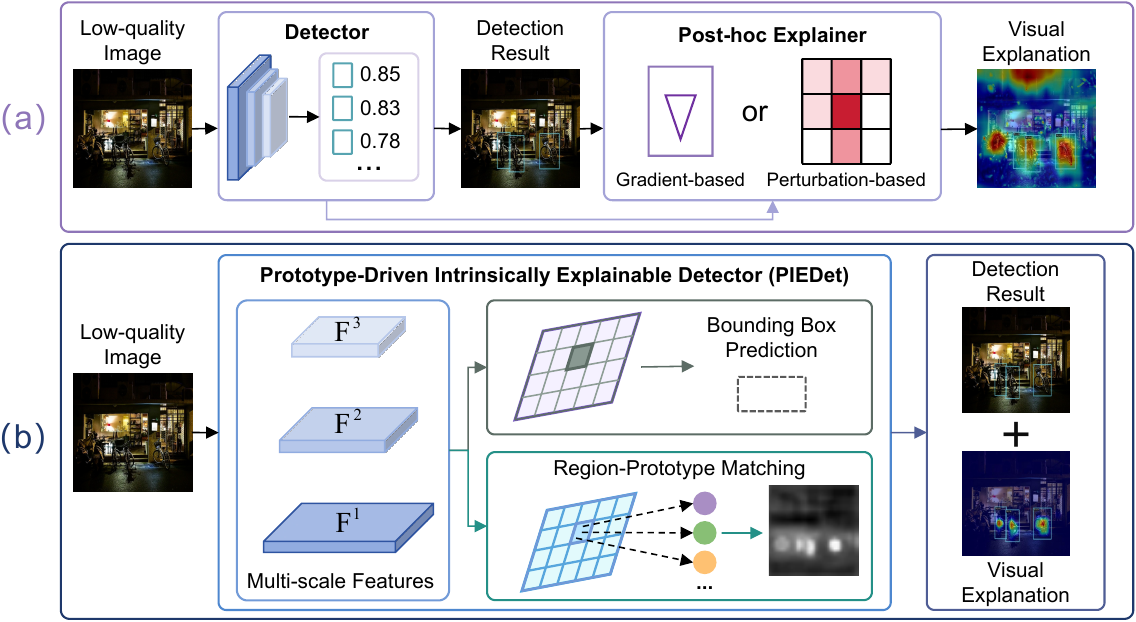}
	\caption{Comparison of explanation mechanisms for object detection. 
		(a) Conventional detector with post-hoc explanation, where visual explanations are generated after detection by an external explainer. 
		(b) Our proposed PIEDet, which integrates prototype matching into detection and directly produces both detection results and intrinsic prototype visual explanation.}
	\label{fig:Architecture_Comparison_Diagram}
\end{figure}

Object detection is a fundamental task in computer vision and has achieved remarkable progress in both detection accuracy and inference speed in recent years~\cite{ref_FasterRCNN, ref_YOLO, ref_DETR}. However, existing object detectors typically make predictions in a black-box manner and output only bounding boxes and category labels, making it difficult to reveal the key visual evidence supporting their predictions. Such decision opacity hinders their trustworthy deployment in safety-critical scenarios, such as autonomous driving and security surveillance~\cite{ref_AD1, ref_AD2}.

To analyze model predictions, researchers commonly apply post-hoc interpretability methods after detection~\cite{ref_CAM, ref_GradCAM, ref_GradCAMpp, ref_ScoreCAM, ref_LayerCAM, ref_FinerCAM, ref_SSGradCAMpp, ref_DRISE, ref_ODAM}, as illustrated by the explanation branch following the detector output in Fig.~\ref{fig:Architecture_Comparison_Diagram}(a). These methods attribute existing predictions using gradients~\cite{ref_GradCAM, ref_GradCAMpp, ref_ODAM} or input perturbations~\cite{ref_DRISE} to estimate which image regions contribute most strongly to the predictions.
However, post-hoc interpretability methods externally attribute prediction results after the detector has completed inference and therefore suffer from two major limitations. On the one hand, additional backward propagation or perturbation sampling increases memory and computational overhead, limiting their applicability in resource-constrained and real-time scenarios. On the other hand, because the attribution procedure does not directly participate in model prediction, the faithfulness of the resulting explanations to the model's internal decision-making process cannot be intrinsically guaranteed.
To overcome these limitations, a more desirable solution is to embed interpretability into the detector's inference process, allowing the model to make category predictions based on explicit, visualizable, and semantically meaningful internal representations while simultaneously generating the corresponding discriminative evidence.

\IEEEpubidadjcol
Prototype learning provides a suitable means of constructing such internal representations. It learns representative semantic prototypes for each class and performs category discrimination by measuring the similarity between region features and class prototypes. Because prototype responses directly participate in prediction and possess explicit class semantics, they can be directly visualized as the model's discriminative evidence, thereby enabling detection results and explanatory information to be generated simultaneously.

Previous studies have demonstrated the semantic modeling capability and interpretability potential of prototype learning in image classification~\cite{ref_ProtoPNet}, semantic segmentation~\cite{ref_segmentation1, ref_segmentation2, ref_segmentation3}, and few-shot object detection~\cite{ref_FSOD1, ref_FSOD2, ref_FSOD3, ref_FSOD4}. However, most existing prototype-based detectors are built upon two-stage frameworks such as Faster R-CNN and perform prototype matching on RoI features after region proposals have been generated. Consequently, their explanations are confined to proposal regions, while the additional region feature extraction process also increases processing complexity.

In contrast, one-stage detectors~\cite{ref_YOLO, ref_CenterNet} directly perform classification and localization at dense spatial locations on multi-scale feature maps, making their prediction paradigm naturally compatible with prototype matching. By formulating category prediction at each spatial location as the similarity between local features and class prototypes, prototype responses can directly form spatial response maps, turning discriminative evidence into an intrinsic output of the classification process. Therefore, the key to integrating prototype learning with one-stage detection lies in constructing scale-aware and discriminative class prototypes that can support dense classification while generating stable and semantically explicit discriminative evidence.

Based on the above analysis, we propose PIEDet, a prototype-driven intrinsically explainable object detection framework, as illustrated in Fig.~\ref{fig:Architecture_Comparison_Diagram}(b). Unlike post-hoc explanation methods that rely on external attribution, PIEDet embeds class prototypes as explicit discriminative units into the classification branches of a one-stage detector. Category predictions are made by matching spatial features with class prototypes, and the corresponding prototype responses are directly visualized as discriminative evidence, enabling detection results and explanatory information to be generated within the same inference process.

To accommodate the multi-instance and multi-scale characteristics of object detection, PIEDet constructs hierarchical class prototypes at detection levels of different scales and introduces prototype regularization, region-to-prototype matching, and scale-aligned hierarchical prototype supervision. These components respectively enhance the inter-class discriminability of the prototypes, guide prototype responses to focus on object regions, and improve the scale specificity and discriminative capability of the hierarchical prototypes. The main contributions of this work are summarized as follows:
\begin{itemize}
	\item We propose a prototype-driven intrinsically explainable object detection framework (PIEDet) that embeds class prototypes as explicit discriminative units into a one-stage detector, allowing detection results and internal discriminative evidence to be generated within the same inference process.
	\item We propose a prototype-driven feature learning method comprising a prototype regularization loss and a region-to-prototype matching loss. The former reduces redundancy among prototypes of different classes and improves inter-class discriminability, while the latter aligns object-region features with class prototypes and encourages prototype responses to semantically focus on object regions.
	\item We propose a scale-aligned hierarchical prototype supervision mechanism. It defines scale-dependent regression ranges for different detection levels and accordingly assigns scale-matched region-level supervision, thereby reducing supervision noise and improving the scale specificity and discriminative capability of the hierarchical prototypes.
	\item Experiments on ExDark, RTTS, and VOC2012-FOG demonstrate that PIEDet improves detection performance while generating intrinsic explanations focused on object regions, and approaches or surpasses mainstream post-hoc interpretability methods in explanation quality.
\end{itemize}

\section{Related Work}
\subsection{Object Detection Methods}
Existing deep-learning-based object detectors can be broadly divided into two-stage detectors, one-stage detectors, and Transformer-based end-to-end detectors. The R-CNN family is representative of two-stage methods. R-CNN~\cite{ref_RCNN} extracts features independently from each proposal and therefore incurs substantial computational cost. SPP-Net~\cite{ref_SPPNet} and Fast R-CNN~\cite{ref_FastRCNN} reduce redundant computation by sharing convolutional features, while Faster R-CNN~\cite{ref_FasterRCNN} further introduces a region proposal network to jointly optimize proposal generation, classification, and regression. These methods generally provide high localization accuracy, but their detection pipelines are relatively complex and inference efficiency is limited.

One-stage detectors remove the explicit proposal generation stage and directly perform object classification and localization. The YOLO family~\cite{ref_YOLO} predicts detections in a single forward pass and has progressively improved the backbone, feature fusion, detection head, and label assignment strategies to achieve a favorable balance between accuracy and efficiency. Recent models represented by YOLOv8 adopt anchor-free prediction and decoupled classification and regression heads. They also incorporate Distribution Focal Loss (DFL)~\cite{ref_DFL}, which models boundary locations as discrete probability distributions to improve localization representation. CornerNet~\cite{ref_CornerNet}, ExtremeNet~\cite{ref_ExtremeNet}, and CenterNet~\cite{ref_CenterNet} represent objects using corners, extreme points, and center points, respectively, reducing their reliance on predefined anchors.

DETR~\cite{ref_DETR} formulates object detection as an end-to-end set prediction problem and uses a Transformer with bipartite matching to establish one-to-one correspondence between object queries and ground-truth instances, eliminating anchors and non-maximum suppression. To address its slow convergence and limited performance on small objects, SAM-DETR~\cite{ref_SAMDETR} improves matching between queries and image features through semantic alignment, whereas Deformable DETR~\cite{ref_DeformableDETR} employs multi-scale deformable attention to accelerate training and improve small-object detection. Nevertheless, these methods are mostly developed for high-quality natural images. Under degradations such as low illumination, haze, low contrast, noise, and color shifts, weakened texture and edge cues can cause feature degradation and foreground--background confusion.

Early studies commonly used low-light enhancement or image dehazing algorithms~\cite{ref_LLIE1, ref_LLIE2, ref_DEHAZING1, ref_DEHAZING2, ref_DEHAZING3, ref_DEHAZING4} as preprocessing for detection. However, the optimization objectives of image restoration and object detection are not fully aligned, and enhancement artifacts or erroneous textures may interfere with recognition and localization. Subsequent work has therefore adopted joint optimization or task-driven feature enhancement. HLA-Face~\cite{ref_HLA-Face} and MAET~\cite{ref_MAET} reduce representation discrepancies between images of different quality through multi-task supervision or feature disentanglement. DSNet~\cite{ref_DSNet} and IA-YOLO~\cite{ref_IA-YOLO} embed learnable enhancement modules into detection frameworks. FeatEnHancer~\cite{ref_FeatEnHancer} and T2~\cite{ref_T2} strengthen object features using illumination information, while YOLA~\cite{ref_YOLA} learns illumination-invariant representations to reduce dependence on explicit enhancement. Although these approaches improve detection under low-quality conditions, they commonly introduce extra branches, auxiliary supervision, or specialized modules, increasing model and training complexity. Moreover, their high-dimensional implicit features still cannot intuitively reveal the semantic information on which detection predictions are based.

\subsection{Explainability Methods for Object Detection}
Research on the interpretability of deep neural networks aims to reveal the discriminative evidence underlying model predictions, thereby improving transparency, trustworthiness, and verifiability. Existing visual explanation methods typically generate saliency maps after the prediction process, highlighting image regions that are considered important for the model's decision.

CAM~\cite{ref_CAM} weights convolutional feature maps using classification-layer weights to obtain class-related spatial responses. Grad-CAM~\cite{ref_GradCAM} and Grad-CAM++~\cite{ref_GradCAMpp} further use gradient information to estimate the contribution of individual feature channels to a target class, making them applicable to a broader range of convolutional architectures. Score-CAM~\cite{ref_ScoreCAM} computes feature-map weights from changes in class confidence during forward inference, reducing dependence on gradient quality. Layer-CAM~\cite{ref_LayerCAM}, Finer-CAM~\cite{ref_FinerCAM}, and SSGrad-CAM++~\cite{ref_SSGradCAMpp} improve saliency maps through local response modeling, fine-grained class discrimination, and gradient stabilization, respectively.
For object detection, D-RISE~\cite{ref_DRISE} randomly masks the input image and observes changes in detection results to estimate the importance of different regions to a particular detection. Although model-agnostic, it requires extensive perturbation sampling and repeated forward inference, resulting in high computational cost. ODAM~\cite{ref_ODAM} extends gradient-weighted explanations to object detection and generates class- and localization-related visualizations for individual detection instances, improving instance specificity.

Although these post-hoc methods provide useful visual evidence, their explanation procedures are separated from the detector's own decision process, making it difficult to guarantee that the generated saliency maps faithfully reflect the model's actual reasoning. In addition, gradient backpropagation, random perturbations, or repeated inference introduce extra computation and memory consumption. More importantly, post-hoc saliency maps indicate where the model attends but generally fail to reveal the concrete semantic concepts represented by those regions.

\begin{figure*}[!t]
	\centering
	\includegraphics[width=\textwidth]{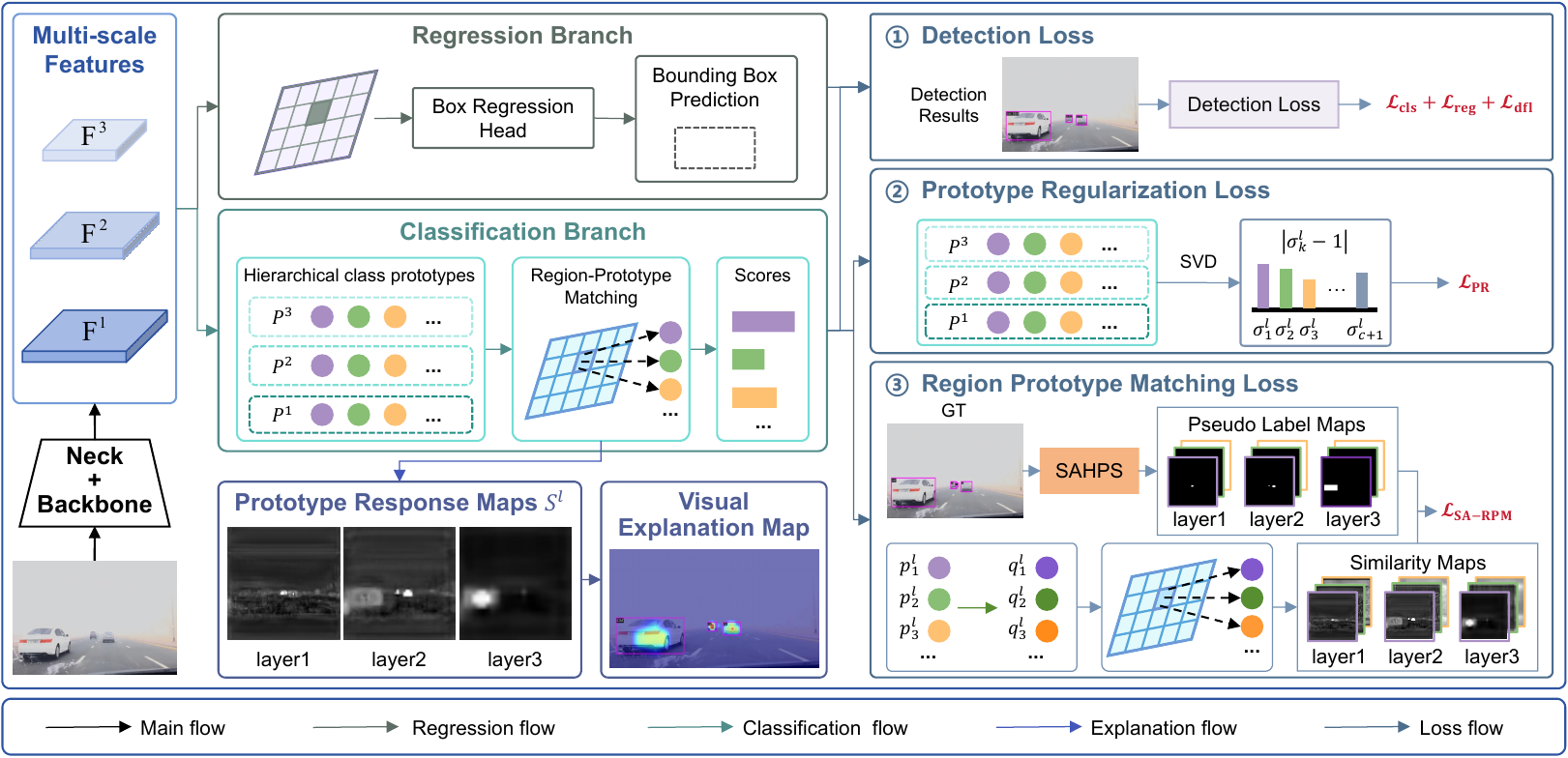}
	\caption{Overview of the PIEDet architecture. PIEDet performs object localization and category prediction using multi-scale features and decoupled detection heads, where the classification branch generates classification responses and intrinsic explanation maps through region-to-prototype matching. During training, the detection loss, prototype regularization loss, and scale-aligned region-to-prototype matching loss are jointly optimized.}
	\label{fig:PIEDet-Architecture}
\end{figure*}

\section{Proposed Method}
\subsection{Overall Detection Architecture}
To improve both the robustness and interpretability of object detection, we propose PIEDet, a prototype-driven intrinsically explainable object detection framework whose overall architecture is shown in Fig.~\ref{fig:PIEDet-Architecture}. The framework performs object localization and classification using multi-scale features and decoupled detection heads. Class prototypes are introduced into the classification branch at every level, transforming category prediction into a matching process between region features and class prototypes. The resulting prototype responses directly participate in category discrimination and can also be fused across scales to generate visual explanation maps, enabling simultaneous output of detections and discriminative evidence. During training, the model jointly optimizes the conventional detection loss, prototype regularization loss, and scale-aligned region-to-prototype matching loss to improve inter-class discriminability in the prototype space and guide different detection levels to focus on objects of corresponding scales. The following sections describe prototype modeling, prototype-driven feature learning, and scale-aligned hierarchical prototype supervision.

\subsection{Intrinsically Explainable Prototype Modeling}
Given an input image $x$, the backbone and feature pyramid network extract a set of multi-scale features:
\begin{equation}
	\mathcal{F}=\{F^l\}_{l=1}^{L}, \quad
	F^l \in \mathbb{R}^{H_l \times W_l \times D_l},
	\label{eq:multi_scale_feature}
\end{equation}
where $l$ indexes the $l$-th feature level, $L$ is the number of feature levels, $H_l$ and $W_l$ denote the spatial dimensions of the corresponding feature map, and $D_l$ is its channel dimension. Because different feature levels have different spatial resolutions and receptive fields, they capture object semantics at different scales.

In the detection head, PIEDet adopts decoupled classification and regression branches. The classification branch learns region-level semantic features for category discrimination, whereas the regression branch predicts object bounding boxes. At the $l$-th detection level, the classification branch first maps the input feature $F^l$ to a region-semantic feature:
\begin{equation}
	Z^l = \phi_{\mathrm{cls}}^l(F^l), \quad
	Z^l \in \mathbb{R}^{H_l \times W_l \times D},
	\label{eq:semantic_feature}
\end{equation}
where $\phi_{\mathrm{cls}}^l(\cdot)$ denotes the feature mapping function in the classification branch at level $l$, and $D$ is the unified dimension of the prototype space. At spatial location $(i,j)$, the region feature vector is denoted by $z_{i,j}^l \in \mathbb{R}^{D}$.

To provide the classification process with an interpretable semantic matching form, PIEDet defines a set of hierarchical class prototypes at every detection level and organizes them into a prototype matrix:
\begin{equation}
	P^l
	=
	[p_1^l, p_2^l, \ldots, p_C^l, p_{\mathrm{bg}}^l]
	\in
	\mathbb{R}^{(C+1)\times D},
	\label{eq:P-l}
\end{equation}
where $C$ is the number of foreground classes, $p_k^l \in \mathbb{R}^{D}$ denotes the semantic prototype of the $k$-th foreground class at the $l$-th detection level, and $p_{\mathrm{bg}}^l$ is the background prototype. Each prototype vector can be interpreted as a discriminative semantic center for one class at the current scale, characterizing its typical feature pattern within the corresponding receptive field.

For the region feature $z_{i,j}^l$ at location $(i,j)$, its matching score and prototype response with respect to the $k$-th class prototype are defined as:
\begin{equation}
	a_{k,i,j}^{l}
	=
	\left\langle z_{i,j}^{l}, \hat{p}_k^l \right\rangle
	+
	b_k^l,
	\qquad
	S_{k,i,j}^{l}
	=
	\sigma\left(a_{k,i,j}^{l}\right),
	\label{eq:prototype_response}
\end{equation}
where $\langle \cdot,\cdot \rangle$ denotes a vector similarity measure, $\hat{p}_k^l$ is the normalized class prototype, $b_k^l$ is the class-specific bias, and $\sigma(\cdot)$ denotes the Sigmoid function. $S_{k,i,j}^{l}$ measures the match between the current location and the $k$-th class prototype. It participates in category discrimination and can also be directly visualized as a class-prototype response map.

Furthermore, PIEDet obtains its intrinsic explanation maps directly from the prototype responses produced during forward inference, without additional backpropagation or perturbation sampling. For a class $k$ to be explained, we upsample the class-prototype response $S_k^l$ from each detection level to the input-image resolution, average the responses, and normalize the result to obtain a class-level explanation map:
\begin{equation}
	E_k=
	\operatorname{Norm}
	\left(
	\frac{1}{L}
	\sum_{l=1}^{L}
	\operatorname{Up}\left(S_k^{l},H,W\right)
	\right),
	\label{eq:prototype_explanation_map}
\end{equation}
where $L$ is the number of detection feature levels, $\operatorname{Up}(\cdot)$ denotes bilinear interpolation, and $\operatorname{Norm}(\cdot)$ denotes Min--Max normalization.

This design gives category prediction in PIEDet an explicit prototype-matching interpretation. In other words, classification is no longer an opaque mapping but is determined by whether a region feature matches a particular class prototype. The feature learning method introduced next further improves the inter-class discriminability and object-region focus of prototype responses, producing clearer, more stable, and semantically consistent explanation maps.

\subsection{Prototype-Driven Feature Learning}
We propose a prototype-driven feature learning method that comprises a prototype regularization loss and a region-to-prototype matching loss. The former reduces redundancy among prototypes of different classes and improves inter-class discriminability, whereas the latter aligns object-region features with class prototypes and encourages prototype responses to semantically focus on object regions.

\subsubsection{Prototype Regularization Loss}
After hierarchical class prototypes have been constructed, ensuring strong inter-class separability among prototype vectors is essential for intrinsically explainable detection. If prototypes of different classes are highly correlated in the feature space, a region feature may activate multiple class prototypes simultaneously, resulting in diffuse prototype responses, ambiguous class boundaries, and less trustworthy explanations. Previous work has shown that orthogonality constraints help disentangle class semantics in the feature space and improve discriminability~\cite{ref_OWN}. We therefore introduce a prototype regularization loss that constrains the geometric relationships among class prototypes from the perspective of the global prototype space.

For the $l$-th detection level, we impose a global geometric constraint directly on the prototype matrix $P^l$ in Eq.~\eqref{eq:P-l}. Conventional methods commonly reduce prototype redundancy using pairwise constraints such as cosine similarity or POP~\cite{ref_POP}. Because these approaches operate mainly on local prototype pairs, they cannot fully characterize the geometry of the entire prototype space. In contrast, we apply a global spectral constraint based on singular value decomposition (SVD) to regularize the prototype matrix as a whole. Specifically, $P^l$ is decomposed as:
\begin{equation}
	P^l
	=
	U^l \Sigma^l {V^l}^{\top},
	\qquad
	\Sigma^l
	=
	\operatorname{diag}
	\left(
	\sigma_1^l,
	\sigma_2^l,
	\ldots,
	\sigma_{C+1}^l
	\right),
	\label{eq:svd}
\end{equation}
where $U^l$ and $V^l$ are the left and right singular-vector matrices, respectively, and $\sigma_k^l$ is the $k$-th singular value. A singular-value-based spectral constraint characterizes the structural stability of the matrix globally and provides a differentiable geometric constraint for optimizing the prototype space~\cite{ref_TDN}. To encourage class prototypes to form a stable and approximately orthogonal discriminative space, we define the prototype regularization loss as:
\begin{equation}
	\mathcal{L}_{\mathrm{PR}}^l
	=
	\sum_{k=1}^{C+1}
	\left|\sigma_k^l-1\right|.
	\label{eq:pr_loss}
\end{equation}

By constraining the singular values to approach 1, this loss keeps the prototype matrix well conditioned and prevents prototypes of different classes from collapsing toward similar directions. It therefore reduces semantic redundancy among class prototypes and improves inter-class separability, providing a more reliable discriminative basis for subsequent region-to-prototype matching.

\subsubsection{Region-to-Prototype Matching Loss}
To establish an explicit correspondence between class prototypes and ground-truth object regions, we propose a region-to-prototype matching loss that provides region-level supervision for prototype responses. Its core principle is that foreground-region features should produce high matching responses with their corresponding class prototypes, whereas background-region features should respond strongly to the background prototype.

Given a ground-truth box $B_n=(x_1^n,y_1^n,x_2^n,y_2^n)$ and its class label $c_n$, the center point in the original image corresponding to location $(i,j)$ at feature level $l$ is:
\begin{equation}
	g_{i,j}^{l}
	=\left(g_{y}^{l},g_{x}^{l}\right)
	=
	\left((i+\frac{1}{2})s_l,
	(j+\frac{1}{2})s_l\right),
	\label{eq:grid_point_basic}
\end{equation}
where $s_l$ denotes the downsampling stride of the $l$-th feature level.

To generate the object-region label map for class $k$ at feature level $l$, we first initialize $Y_k^l \in \mathbb{R}^{H_l \times W_l}$ as an all-zero matrix, where $Y_{k,i,j}^l$ indicates whether location $(i,j)$ at level $l$ belongs to an object region of class $k$. For a ground-truth box $B_n$ of class $k$, a location is considered part of the valid interior of that object if its corresponding center point in the original image lies inside the box and remains half a stride away from every box boundary. Specifically, the spatial relationship between the center point corresponding to $(i,j)$ and the ground-truth box $B_n$ is constrained as follows:
\begin{equation}
	x_1^n+\frac{s_l}{2}
	\le g_x^l \le
	x_2^n-\frac{s_l}{2},
	\qquad
	y_1^n+\frac{s_l}{2}
	\le g_y^l \le
	y_2^n-\frac{s_l}{2}.
	\label{eq:positive_region_basic}
\end{equation}

When both conditions in Eq.~\eqref{eq:positive_region_basic} are satisfied, the corresponding location in $Y_k^l$ is marked as positive, i.e., $Y_k^l(i,j)=1$; otherwise it remains 0. This constraint approximately ensures that the local region represented by the feature point lies predominantly inside the object box, avoiding uncertain supervision caused by mixed foreground and background features near box boundaries. For multiple ground-truth boxes of the same class, their union is used to update the regional label map. Therefore, the region-matching label maps at level $l$ are:
\begin{equation}
	Y^l
	=
	\{Y_1^l,Y_2^l,\ldots,Y_C^l,Y_{\mathrm{bg}}^l\},
	\quad
	Y_k^l \in \{0,1\}^{H_l \times W_l},
	\label{eq:basic_label_map}
\end{equation}
where $Y_k^l$ denotes the object-region label map of class $k$. The background label map is defined as the complement of the union of all foreground-class regions:
\begin{equation}
	Y_{\mathrm{bg}}^l
	=
	1-\max_{k=1}^{C}Y_k^l.
	\label{eq:basic_bg_label}
\end{equation}

After obtaining the region-matching label maps, PIEDet computes matching responses between region features and class prototypes. Because region features of one class may be interfered with by prototypes of other classes, we further introduce the concept of an orthogonal subspace to improve the class specificity of prototype responses. For the $k$-th class prototype at feature level $l$, we denote the target prototype as $d_k^l=p_k^l$ and construct an interference-prototype set from all prototypes except that class:
\begin{equation}
	U_k^l
	=
	[p_1^l,\ldots,p_{k-1}^l,p_{k+1}^l,\ldots,p_C^l,p_{\mathrm{bg}}^l]
	\in
	\mathbb{R}^{(C)\times D}.
	\label{eq:interference_proto}
\end{equation}

The corresponding orthogonal-subspace prototype matching operator is defined as:
\begin{equation}
	\Pi_k^l
	=
	I-U_k^l{U_k^l}^{\dagger},
	\qquad
	q_k^l
	=
	\frac{\Pi_k^l d_k^l}
	{{d_k^l}^{\top}\Pi_k^l d_k^l+\epsilon},
	\label{eq:osp_operator}
\end{equation}
where ${U_k^l}^{\dagger}$ is the pseudoinverse of $U_k^l$, $I$ is the identity matrix, and $\epsilon$ is a numerical-stability term. For the region feature $z_{i,j}^l$ at location $(i,j)$, its orthogonal-subspace prototype response is:
\begin{equation}
	\widetilde{S}_{k,i,j}^{l}
	=
	\sigma \left\langle {z_{i,j}^{l}}^{\top}, q_k^l \right\rangle,
	\label{eq:osp_response}
\end{equation}
where $\widetilde{S}_{k,i,j}^{l}$ is the matching response between the region feature at location $(i,j)$ and the class-$k$ prototype after interference from other class prototypes has been suppressed. Unlike Eq.~\eqref{eq:prototype_response}, Eq.~\eqref{eq:osp_response} is used primarily for region-to-prototype matching supervision and therefore does not include an additional class bias. The class biases are learned under the detection classification loss.

Finally, given the prototype response maps $\widetilde{S}^{l}$ at feature level $l$ and the constructed region label maps $Y^{l}$, we apply binary cross-entropy to supervise prototype responses for all classes at the region level. Let $\mathcal{K}=\{1,2,\ldots,C,\mathrm{bg}\}$ denote the set containing all foreground classes and the background class. The region-to-prototype matching loss at feature level $l$ is defined as:
\begin{equation}
	\label{eq:rpm_loss}
	\begin{aligned}
		\mathcal{L}_{\mathrm{RPM}}^{l}
		= {}&
		-\frac{1}{|\mathcal{K}| \times H_l \times W_l}
		\sum_{k\in\mathcal{K}}
		\Big[
		Y_k^l \cdot
		\log\left(\widetilde{S}_k^l+\epsilon\right)
		\\
		&\quad+
		\left(1-Y_k^l\right)
		\cdot
		\log\left(1-\widetilde{S}_k^l+\epsilon\right)
		\Big],
	\end{aligned}
\end{equation}
where $\epsilon$ is a small constant used to prevent numerical instability.

\subsection{Scale-aligned Hierarchical Prototype Supervision}
The region-to-prototype matching loss establishes a correspondence between object regions and class prototypes. However, using all ground-truth boxes to supervise every feature level ignores differences in spatial resolution and receptive field among detection levels. This introduces scale-mismatched supervision noise and weakens the scale specificity and discriminability of hierarchical prototypes.

To address this problem, we propose a scale-aligned hierarchical prototype supervision mechanism in which class prototypes at each feature level receive supervision only from object regions that match their effective perception range. Specifically, PIEDet incorporates the scale prediction range of the regression branch into prototype supervision. Following the discrete-distribution regression formulation of Distribution Focal Loss (DFL)~\cite{ref_DFL}, we define a scale-dependent regression range for the $l$-th feature level:
\begin{equation}
	\mathcal{R}_l
	=
	[0,\tau_l s_l],
	\label{eq}
\end{equation}
where $s_l$ denotes the downsampling stride of the $l$-th feature level, and $\tau_l$ is a scale-control coefficient that determines the upper bound of its regression space.

Given a ground-truth box $B_n=(x_1^n,y_1^n,x_2^n,y_2^n)$, its width and height are $w_n=x_2^n-x_1^n$ and $h_n=y_2^n-y_1^n$, respectively. If the object size satisfies $w_n, h_n \in \mathcal{R}_l$, the object is regarded as a valid supervision target for feature level $l$. PIEDet retains these scale-matched ground-truth bounding boxes and constructs the scale-aligned region label map $Y_{\mathrm{SA}}^l$ following the region-label generation procedure described in Eqs.~\eqref{eq:positive_region_basic}--\eqref{eq:basic_bg_label}.

During training, we replace the basic region label map $Y^l$ in Eq.~\eqref{eq:rpm_loss} with $Y_{\mathrm{SA}}^l$, yielding the scale-aligned region-to-prototype matching loss:
\begin{equation}
	\mathcal{L}_{\mathrm{SA\mbox{-}RPM}}^l
	=
	\mathcal{L}_{\mathrm{RPM}}^l
	\left(
	\widetilde{S}^{l},
	Y_{\mathrm{SA}}^l
	\right).
	\label{eq:sa_rpm_loss}
\end{equation}

With this mechanism, class prototypes at different feature levels learn object semantics matched to their respective receptive fields. It reduces prototype confusion caused by cross-scale supervision noise and consequently improves the robustness of multi-scale detection and the reliability of explanations.

\subsection{Training Objective}
The overall training objective of PIEDet consists of the conventional detection losses, the prototype regularization loss, and the scale-aligned region-to-prototype matching loss. The conventional detection losses ensure accurate category prediction and bounding-box localization, the prototype regularization loss constrains redundancy among class prototypes, and the scale-aligned region-to-prototype matching loss guides prototype responses at different detection levels to focus on object regions matched to their receptive fields. The complete objective is defined as:
\begin{equation}
	\mathcal{L}_{\mathrm{total}}
	=
	\mathcal{L}_{\mathrm{cls}}
	+
	\mathcal{L}_{\mathrm{reg}}
	+
	\mathcal{L}_{\mathrm{dfl}}
	+
	\sum_{l=1}^{L}
	\left(
	\mathcal{L}_{\mathrm{PR}}^l
	+
	\mathcal{L}_{\mathrm{SA\mbox{-}RPM}}^l
	\right),
	\label{eq}
\end{equation}
where $L$ is the number of detection feature levels, and $\mathcal{L}_{\mathrm{cls}}$, $\mathcal{L}_{\mathrm{reg}}$, and $\mathcal{L}_{\mathrm{dfl}}$ denote the classification loss, bounding-box regression loss, and Distribution Focal Loss, respectively.

\begin{table*}[!t]
	\caption{Quantitative comparison of our proposed PIEDet with other representative methods on low-light scenes. All results are reported in percentage (\%). The best results are highlighted in \textbf{bold}, and the second-best in \underline{underline}. $^{\dagger}$Results of T2 are reported as in the original paper. $^{\ddagger}$YOLA is re-implemented with YOLOv8 trained from scratch.}
	\label{tab:low-light_comparison}
	\centering
	\renewcommand{\arraystretch}{1.05}
	\resizebox{\textwidth}{!}{
		\begin{tabular}{@{\hspace{0.8em}}
				>{\centering\arraybackslash}m{3.1cm}
				>{\centering\arraybackslash}m{2.8cm} |
				>{\centering\arraybackslash}m{1.0cm}
				>{\centering\arraybackslash}m{0.6cm}
				>{\centering\arraybackslash}m{0.9cm}
				>{\centering\arraybackslash}m{0.5cm}
				>{\centering\arraybackslash}m{0.5cm}
				>{\centering\arraybackslash}m{0.5cm}
				>{\centering\arraybackslash}m{0.8cm}
				>{\centering\arraybackslash}m{0.5cm}
				>{\centering\arraybackslash}m{0.5cm}
				>{\centering\arraybackslash}m{1.3cm}
				>{\centering\arraybackslash}m{1.0cm}
				>{\centering\arraybackslash}m{0.8cm} |
				>{\centering\arraybackslash}m{1.3cm}@{\hspace{0.8em}}}
			\hline
			{Enhancer} & {Detector} & Bicycle & Boat & Bottle & Bus & Car & Cat & Chair & Cup & Dog & Motorbike & People & Table & mAP@0.5 \\
			\hline
			-- & YOLOv8 & 71.8 & 64.2 & 69.2 & 82.8 & 76.8 & 51.7 & 54.4 & 58.9 & 62.5 & 61.9 & 69.6 & 48.4 & 64.4 \\
			MBLLEN~\cite{ref_MBLLEN} & YOLOv8 & 70.8 & 66.7 & 71.0 & 84.1 & 78.0 & 50.1 & 53.9 & 60.3 & 62.6 & 60.7 & 71.0 & 52.8 & 65.2 \\
			KinD~\cite{ref_KinD} & YOLOv8 & 66.5 & 62.9 & 60.0 & 79.1 & 72.1 & 47.5 & 49.1 & 48.5 & 56.0 & 53.8 & 62.6 & 45.6 & 58.7 \\
			Zero-DCE~\cite{ref_Zero-DCE} & YOLOv8 & 71.5 & 68.2 & 70.6 & 85.5 & 79.2 & 51.9 & 55.1 & \underline{64.1} & 63.8 & 59.5 & 72.3 & 50.6 & 66.0 \\
			SCI~\cite{ref_SCI} & YOLOv8 & 71.1 & 67.8 & 71.2 & 83.7 & 79.0 & 52.6 & 55.1 & 62.9 & 61.1 & 60.3 & \underline{72.6} & 51.6 & 65.8 \\
			Retinexformer~\cite{ref_Retinexformer} & YOLOv8 & \textbf{78.5} & \textbf{71.3} & \underline{72.7} & 83.5 & 76.3 & 57.8 & 53.2 & 58.2 & \textbf{69.1} & 63.1 & 70.6 & 47.9 & \underline{66.8} \\
			DiffUIR~\cite{ref_DiffUIR} & YOLOv8 & \underline{77.3} & 69.4 & 65.4 & 81.2 & 73.7 & 56.4 & 54.2 & 57.5 & 67.4 & 62.1 & 70.4 & 48.7 & 65.3 \\
			Diff-Retinex++~\cite{ref_Diff-Retinex++} & YOLOv8 & 72.6 & 65.2 & 70.2 & 83.4 & 77.7 & 52.7 & 54.3 & 60.7 & 63.6 & 61.2 & 69.9 & 51.2 & 65.2 \\
			-- & T2$^{\dagger}$~\cite{ref_T2} & 63.5 & 57.9 & 41.9 & \textbf{88.2} & 67.1 & \textbf{64.2} & 51.9 & 42.9 & 60.4 & \underline{63.8} & 58.7 & \textbf{56.5} & 59.7 \\
			-- & FeatEnHancer~\cite{ref_FeatEnHancer} & 72.1 & 67.4 & 70.3 & 84.4 & \underline{79.5} & 55.8 & \underline{57.9} & 62.3 & 64.6 & 61.1 & 72.1 & 51.2 & 66.5 \\
			-- & YOLA$^{\ddagger}$~\cite{ref_YOLA} & 66.0 & 55.9 & 64.1 & 79.0 & 73.1 & 42.4 & 45.1 & 55.9 & 48.5 & 49.1 & 65.2 & 38.9 & 56.9 \\
			\hline
			-- & PIEDet (ours) & 74.1 & \underline{70.5} & \textbf{74.5} & \underline{87.3} & \textbf{80.4} & \underline{59.7} & \textbf{58.6} & \textbf{65.1} & \underline{68.0} & \textbf{64.2} & \textbf{73.0} & \underline{53.8} & \textbf{69.1} \\
			\hline
		\end{tabular}
	}
\end{table*}

\begin{table*}[!t]
	\caption{Complementary quantitative comparison on low-light scenes using COCO-style evaluation metrics. All results are reported in percentage (\%). The best results are highlighted in \textbf{bold}, and the second-best in \underline{underline}.}
	\label{tab:complementary-quantitative-comparison}
	\centering
	\renewcommand{\arraystretch}{1.05}
	\resizebox{\textwidth}{!}{
		\begin{tabular}{@{\hspace{0.8em}}
				>{\centering\arraybackslash}m{3.1cm}
				>{\centering\arraybackslash}m{2.8cm} |
				>{\centering\arraybackslash}m{2.8cm}
				>{\centering\arraybackslash}m{2.2cm}
				>{\centering\arraybackslash}m{2.2cm}
				>{\centering\arraybackslash}m{2.0cm}
				>{\centering\arraybackslash}m{2.0cm}
				>{\centering\arraybackslash}m{2.0cm}
				@{\hspace{0.8em}}}
			\hline
			{Enhancer} & {Detector} & {mAP@0.50:0.95} & {mAP@0.5} & {mAP@0.75} & {mAP$_s$} & {mAP$_m$} & {mAP$_l$} \\
			\hline
			-- & YOLOv8 & 39.7 & 64.4 & 43.4 & 10.9 & 31.0 & 45.3 \\
			Retinexformer~\cite{ref_Retinexformer} & YOLOv8 & 41.3 & \underline{66.8} & 45.0 & \textbf{12.6} & 30.7 & \underline{47.8} \\
			DiffUIR~\cite{ref_DiffUIR} & YOLOv8 & 40.4 & 65.3 & 44.1 & 10.0 & 30.5 & 46.5 \\
			Diff-Retinex++~\cite{ref_Diff-Retinex++} & YOLOv8 & 40.7 & 65.2 & 44.8 & \underline{12.3} & 30.6 & 46.9 \\
			-- & FeatEnHancer~\cite{ref_FeatEnHancer} & \underline{41.7} & 66.5 & \underline{45.8} & 11.0 & \underline{31.8} & \underline{47.8} \\
			-- & YOLA~\cite{ref_YOLA} & 34.0 & 56.9 & 36.6 & 10.4 & 28.7 & 37.8 \\
			\hline
			-- & PIEDet (ours) & \textbf{43.2} & \textbf{69.1} & \textbf{47.8} & 11.7 & \textbf{33.6} & \textbf{49.2} \\
			\hline
		\end{tabular}
	}
\end{table*}

\begin{table}[!t]
	\caption{Quantitative comparison (mAP@0.5) of our proposed PIEDet with other representative methods on hazy scenes.}
	\label{tab:hazy_comparison}
	\centering
	\renewcommand{\arraystretch}{1.05}
	\resizebox{\columnwidth}{!}{
		\begin{tabular}{@{\hspace{0.8em}}
				c
				c |
				>{\centering\arraybackslash}m{2.0cm}
				>{\centering\arraybackslash}m{3.0cm}
				@{\hspace{0.8em}}}
			\hline
			{Enhancer} & {Detector} & {RTTS} & {VOC2012-FOG} \\
			\hline
			-- & YOLOv8 & 74.4 & 58.9 \\
			DNMGDT~\cite{ref_DNMGDT} & YOLOv8 & \underline{74.6} & 57.6 \\
			DiffDehaze~\cite{ref_DiffDehaze} & YOLOv8 & 63.0 & 57.9 \\
			SGDN~\cite{ref_SGDN} & YOLOv8 & 73.9 & 57.7 \\
			-- & MASFNet~\cite{ref_MASFNet} & 67.1 & \underline{59.2} \\
			\hline
			-- & PIEDet (ours) & \textbf{76.0} & \textbf{63.7} \\
			\hline
		\end{tabular}
	}
\end{table}

\section{Experiments}
\subsection{Evaluation Metrics}
To comprehensively evaluate PIEDet, we analyze both detection performance and interpretability. Detection performance is evaluated using COCO-style metrics, including $\mathrm{mAP}@0.50{:}0.95$, $\mathrm{mAP}@0.50$, $\mathrm{mAP}@0.75$, $\mathrm{mAP}_s$, $\mathrm{mAP}_m$, and $\mathrm{mAP}_l$. We also report FPS, FLOPs, and the number of parameters to assess inference efficiency, computational complexity, and model size.

For interpretability, we evaluate faithfulness, spatial consistency, and explanation efficiency. First, insertion and deletion experiments measure the faithfulness of explanation maps to model predictions. Ins. Class and Del. Class assess explanation quality in terms of category confidence, whereas Ins. IoU and Del. IoU measure the influence of explanations on bounding-box localization consistency. Second, Point Game and Energy Point Game evaluate spatial consistency between explanation maps and ground-truth object regions. Point Game checks whether the maximum-response point lies inside the object region, while Energy Point Game measures the proportion of explanation energy distributed within the object region. Comp. measures the compactness of an explanation map, where a lower value indicates a more concentrated response. Finally, Time and Memory quantify the runtime and GPU memory consumption required to generate explanations. Lower Time and Memory indicate greater computational efficiency and deployment friendliness.

Because PIEDet generates class-level explanation maps during inference rather than instance-level maps for individual detections, we adopt a class-level joint explanation protocol for insertion and deletion evaluation. Specifically, for a given class in an image, we explain only high-quality detections: the predicted category must be correct, and both the classification confidence and the IoU between the predicted box and its matched ground-truth box must exceed 0.5. For all qualifying instances of that class, the class-specific explanation map is used to progressively insert or delete image regions, while changes in confidence and localization are recorded throughout the perturbation process. The changes are then averaged over all explained instances of the same class to obtain the class-level explanation score for the current image. This protocol is consistent with PIEDet's intrinsic class-level explanation mechanism and reflects how effectively one class-level map explains multiple instances of the same category.

\subsection{Datasets}
To validate the detection performance and interpretability of our method, we conduct experiments on three challenging low-quality object detection datasets: Exclusively Dark (ExDark)~\cite{ref_exdark}, the Real-world Task-Driven Testing Set (RTTS)~\cite{ref_rtts}, and VOC2012-FOG.

\textbf{ExDark} is a low-light object detection dataset containing 7,363 real low-light images from 12 object categories. We randomly split it into 5,896 training images and 1,467 test images.

\textbf{RTTS} is a real-world hazy-scene dataset from the RESIDE benchmark. It contains 4,322 images covering five categories: bicycle, bus, car, motorbike, and person. We use 3,889 images for training and 433 images for testing.

\textbf{VOC2012-FOG} is a synthetic hazy dataset constructed from Pascal VOC2012~\cite{ref_voc}. We generate hazy images using the atmospheric scattering model:
\begin{equation}
	I(x) = J(x)\cdot e^{-\beta d(x)} + A(1 - e^{-\beta d(x)}),
\end{equation}
where $J(x)$ denotes the clear image, $A$ is the global atmospheric light, $\beta$ is the scattering coefficient, and $d(x)$ is defined as
\begin{equation}
	d(x) = -0.04\cdot\rho + \sqrt{\max(\text{row}, \text{col})}.
\end{equation}

In our implementation, we set $A=0.5$ and $\beta=0.1$ to simulate moderate haze. The hazy images are synthesized offline to avoid additional overhead during training while retaining the original annotations. The dataset contains 11,540 images in total, of which 9,232 are used for training and 2,308 for testing.

\subsection{Implementation Details}
All experiments are conducted on a server equipped with eight NVIDIA GeForce RTX 3090 GPUs using Python 3.9 and PyTorch 2.1.0. For reproducibility, the random seed is set to 5. The model is trained for 100 epochs using stochastic gradient descent (SGD) with a momentum of 0.937, a weight decay of 0.0005, and an initial learning rate of 0.01. For VOC2012-FOG, input images are resized to $512 \times 512$ and the batch size is set to 16. For RTTS and ExDark, images are resized to $800 \times 800$ and the batch size is set to 8 to accommodate the greater memory consumption of higher-resolution inputs. According to the ablation analysis in Sec.~\ref{sec_AS-Modules}, the hyperparameters $\tau_1$, $\tau_2$, and $\tau_3$ are set to 4, 8, and $H/s_3$, respectively, where $H$ denotes the input image height and $s_3$ is the downsampling stride of the final feature level. Dot-product similarity is used for all matching operations between region features and class prototypes.

\subsection{Detection Performance Comparison}
\subsubsection{Low-light Scenes}
We first evaluate the low-light detection performance of PIEDet on ExDark and compare it with two-stage enhancement-based detection methods and end-to-end detection methods. The two-stage methods include MBLLEN~\cite{ref_MBLLEN}, KinD~\cite{ref_KinD}, Zero-DCE~\cite{ref_Zero-DCE}, SCI~\cite{ref_SCI}, Retinexformer~\cite{ref_Retinexformer}, DiffUIR~\cite{ref_DiffUIR}, and Diff-Retinex++~\cite{ref_Diff-Retinex++}; they first enhance the low-light images and then feed the enhanced results to a detector. The end-to-end methods include T2~\cite{ref_T2}, FeatEnHancer~\cite{ref_FeatEnHancer}, and YOLA~\cite{ref_YOLA}. In our experiments, FeatEnHancer is implemented on YOLOv8.

The quantitative results are reported in Table~\ref{tab:low-light_comparison}. PIEDet achieves the highest mAP@0.5 of 69.1\%, outperforming YOLOv8 and the second-best method, Retinexformer, by 4.7 and 2.3 percentage points, respectively. PIEDet also achieves the best results for Bottle, Car, Chair, Cup, Motorbike, and People and ranks second for several other categories, demonstrating stronger class discrimination and robustness under low-light conditions.

Table~\ref{tab:complementary-quantitative-comparison} further reports the COCO-style evaluation results. PIEDet ranks first in mAP, mAP@0.5, and mAP@0.75, reaching 43.2\%, 69.1\%, and 47.8\%, respectively, which confirms its consistent advantage across different IoU thresholds. Across object scales, PIEDet obtains the best performance on medium and large objects and still outperforms YOLOv8 on small objects. Overall, these results show that PIEDet effectively improves low-light object detection without introducing an additional image enhancement module.

\begin{figure*}[!t]
	\centering
	\includegraphics[width=\textwidth]{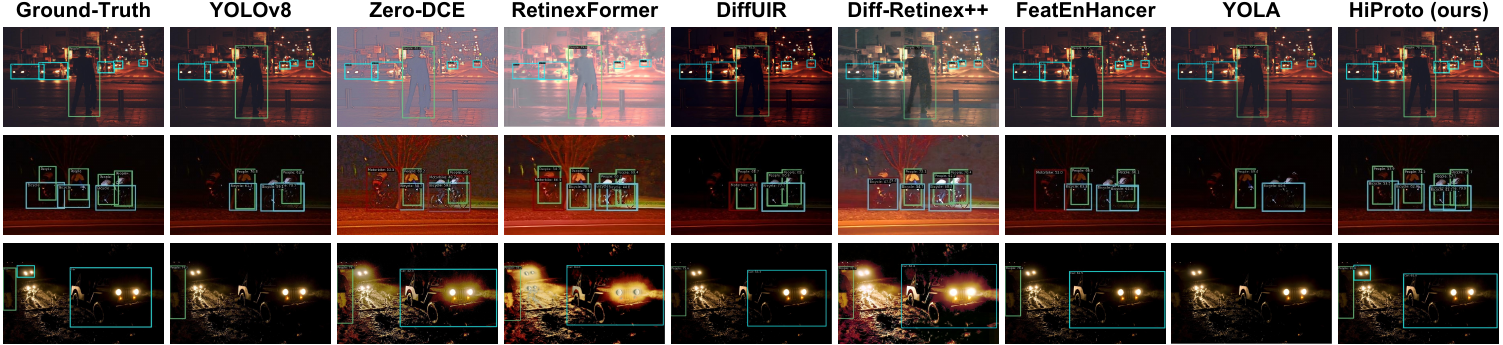}
	\caption{Visual comparison of detection results produced by different methods in low-light scenes.}
	\label{fig:Low-light-detection-results}
\end{figure*}

Figure~\ref{fig:Low-light-detection-results} presents detection results on ExDark. Under low illumination, several competing methods suffer from missed detections or false positives, particularly for poorly illuminated, blurry-edged, or small objects. In contrast, PIEDet localizes objects more consistently while maintaining high category confidence, indicating that hierarchical class prototypes strengthen semantic perception of weak visual cues.

\subsubsection{Hazy Scenes}
We further evaluate PIEDet under hazy conditions on RTTS and VOC2012-FOG. The competing methods include the two-stage dehazing-preprocessing approaches DNMGDT~\cite{ref_DNMGDT}, DiffDehaze~\cite{ref_DiffDehaze}, and SGDN~\cite{ref_SGDN}, as well as the end-to-end method MASFNet~\cite{ref_MASFNet}.

As shown in Table~\ref{tab:hazy_comparison}, PIEDet achieves the highest detection accuracy on both hazy datasets. On RTTS, it improves over YOLOv8 by 1.6 percentage points and over the strongest competing method, DNMGDT, by 1.4 percentage points. On VOC2012-FOG, it outperforms YOLOv8 by 4.8 percentage points and MASFNet by 4.5 percentage points. These results indicate that, even when haze blurs object boundaries and increases background interference, PIEDet obtains more stable detections through explicit matching between region features and class prototypes.

\begin{figure*}[!t]
	\centering
	\includegraphics[width=\textwidth]{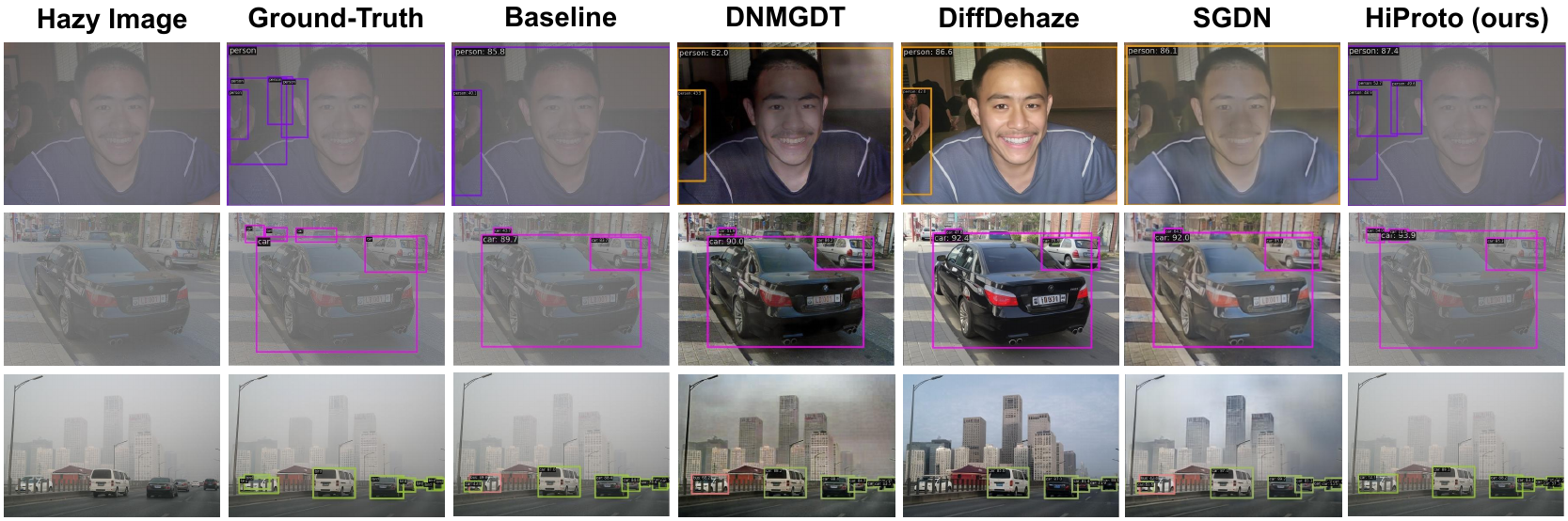}
	\caption{Visual comparison of detection results produced by different methods in hazy scenes.}
	\label{fig:SM-hazy-detection}
\end{figure*}

Figure~\ref{fig:SM-hazy-detection} shows detection results on RTTS and VOC2012-FOG. In hazy scenes, degraded boundaries and texture details make it difficult for detectors to distinguish foreground objects from the background. PIEDet provides more stable detections in these conditions and reduces false positives and missed detections caused by haze, further demonstrating the robustness of prototype-driven feature learning in complex scenes.

\begin{table}[!t]
	\caption{Comparison of different detection methods under normal illumination on the VOC2012 dataset.}
	\label{tab:voc2012_comparison}
	\centering
	\renewcommand{\arraystretch}{1.05}
	\setlength{\tabcolsep}{6pt}
	\resizebox{\columnwidth}{!}{
		\begin{tabular}{
				>{\centering\arraybackslash}m{2.8cm} |
				>{\centering\arraybackslash}m{2.5cm}
				>{\centering\arraybackslash}m{1.8cm}
				>{\centering\arraybackslash}m{1.8cm}
			}
			\hline
			{Method} & {mAP@0.50:0.95} & {mAP@0.5} & {mAP@0.75} \\
			\hline
			YOLOv8 & \underline{0.444} & 0.623 & \underline{0.476} \\
			FeatEnHancer~\cite{ref_FeatEnHancer} & 0.438 & 0.615 & 0.475 \\
			YOLA~\cite{ref_YOLA} & 0.342 & 0.518 & 0.373 \\
			MASFNet~\cite{ref_MASFNet} & -- & \underline{0.629} & -- \\
			\hline
			PIEDet (Ours) & \textbf{0.475} & \textbf{0.663} & \textbf{0.516} \\
			\hline
		\end{tabular}
	}
\end{table}

\begin{table}[!t]
	\caption{Comparison of efficiency metrics among the YOLOv8, our method, and other end-to-end detectors.}
	\label{tab:efficiency_metrics}
	\centering
	\renewcommand{\arraystretch}{1.05}
	\resizebox{\columnwidth}{!}{
		\begin{tabular}{
				>{\centering\arraybackslash}m{3.0cm} |
				>{\centering\arraybackslash}m{1.5cm}
				>{\centering\arraybackslash}m{2.0cm}
				>{\centering\arraybackslash}m{2.5cm}
			}
			\hline
			{Method} & FPS $\uparrow$ & FLOPs (G) $\downarrow$ & Params (M) $\downarrow$ \\
			\hline
			YOLOv8        & \underline{132.67} & \underline{22.317} & 11.164 \\
			FeatEnHancer~\cite{ref_FeatEnHancer}    & 58.28 & 22.323 & 11.303 \\
			YOLA~\cite{ref_YOLA}    		& 80.74 & 28.087 & 11.172 \\
			MASFNet~\cite{ref_MASFNet}    		& 125.00 & 39.616 & \textbf{6.034} \\
			\hline
			PIEDet (Ours)  & \textbf{137.23} & \textbf{21.266} & \underline{10.815} \\
			\hline
		\end{tabular}
	}
\end{table}

\subsubsection{Normal Illumination}
To verify that PIEDet's performance gains are not limited to low-quality imaging conditions, we further conduct experiments on the normally illuminated VOC2012 dataset, as reported in Table~\ref{tab:voc2012_comparison}. PIEDet achieves the best results for mAP, mAP@0.5, and mAP@0.75. Its mAP@0.5 reaches 0.663, exceeding YOLOv8 by 0.040 and the strongest comparable method, MASFNet, by 0.034. It also obtains an mAP of 0.475 and an mAP@0.75 of 0.516, improving over YOLOv8 by 0.031 and 0.040, respectively. These results demonstrate that hierarchical prototype modeling in PIEDet is not restricted to low-light or hazy imagery and provides strong general-purpose detection performance.

\subsubsection{Efficiency Analysis}
Table~\ref{tab:efficiency_metrics} compares PIEDet with other end-to-end detectors in inference speed, computational cost, and parameter count. PIEDet achieves the highest inference speed of 137.23 FPS and the lowest FLOPs of 21.266G among all compared methods. Its parameter count remains close to that of YOLOv8, indicating that the proposed hierarchical prototype mechanism introduces no substantial additional computational burden. These results also show that PIEDet can generate intrinsic explanatory evidence while preserving real-time inference capability.

\begin{table*}[!t]
	\caption{Quantitative comparison between PIEDet and mainstream post hoc interpretability methods on ExDark, RTTS, and VOC2012-FOG datasets. Time is measured in seconds (s), and Memory is measured in megabytes (MB).}
	\label{tab:explainability_quantitative}
	\centering
	\scriptsize
	\setlength{\tabcolsep}{2.6pt}
	\renewcommand{\arraystretch}{1.05}
	\resizebox{\textwidth}{!}{
		\begin{tabular}{@{\hspace{0.8em}}ll|cccc|ccc|cc@{{\hspace{0.8em}}}}
			\hline
			{Dataset} & {Method}
			& Ins. Class $\uparrow$ & Del. Class $\downarrow$
			& Ins. IoU $\uparrow$ & Del. IoU $\downarrow$
			& Point Game $\uparrow$ & Energy PG $\uparrow$
			& Comp. $\downarrow$ & Time $\downarrow$ & Memory $\downarrow$ \\
			\hline
			
			\multirow{6}{*}{ExDark}
			& GradCAM
			& 0.0755 & 0.5766 & 0.1194 & 0.3346 & 0.2699 & 0.1529 & 2.7177 & \underline{1845.5072} & \underline{9353} \\
			& GradCAM++
			& 0.1074 & 0.5211 & 0.1436 & 0.3234 & 0.2718 & 0.1509 & 2.7855 & 1964.5813 & \underline{9353} \\
			& ScoreCAM
			& 0.1070 & 0.4694 & 0.1079 & 0.3031 & 0.4049 & \textbf{0.3972} & \textbf{0.2134} & 11062.1576 & \textbf{999} \\
			& DRISE
			& 0.0702 & 0.3937 & 0.1489 & 0.2982 & \textbf{0.4620} & 0.1269 & 2.9304 & 38768.7614 & \textbf{999} \\
			& ODAM
			& \textbf{0.1912} & \textbf{0.3566} & \textbf{0.1910} & \textbf{0.2608} & \underline{0.4117} & 0.3555 & 0.9479 & 2043.8341 & \underline{9353} \\
			& PIEDet (ours)
			& \underline{0.1621} & \underline{0.3791} & \underline{0.1628} & \underline{0.2851} & 0.3994 & \underline{0.3955} & \underline{0.8856} & \textbf{1313.8374} & \textbf{999} \\
			
			\hline
			
			\multirow{6}{*}{RTTS}
			& GradCAM
			& 0.0385 & 0.3265 & 0.0766 & 0.1726 & 0.1631 & 0.0884 & 3.7642 & \underline{1363.6405} & \underline{9345} \\
			& GradCAM++
			& 0.0247 & 0.3659 & 0.0711 & 0.1833 & 0.1617 & 0.0779 & 3.8077 & 1683.9527 & \underline{9345} \\
			& ScoreCAM
			& 0.0322 & 0.3549 & 0.0537 & 0.1763 & 0.1574 & 0.1476 & \textbf{0.1064} & 7014.2718 & \textbf{996} \\
			& DRISE
			& 0.0658 & \underline{0.2784} & \underline{0.1048} & 0.1704 & \textbf{0.3734} & 0.0733 & 4.1832 & 15100.4041 & \textbf{996} \\
			& ODAM
			& \underline{0.0900} & \textbf{0.2709} & 0.1005 & \textbf{0.1523} & \underline{0.1845} & \underline{0.1740} & \underline{1.1221} & 1790.1539 & \underline{9345} \\
			& PIEDet (ours)
			& \textbf{0.1005} & 0.2793 & \textbf{0.1066} & \underline{0.1618} & 0.1602 & \textbf{0.1794} & 1.2533 & \textbf{707.9164} & \textbf{996} \\
			
			\hline
			
			\multirow{6}{*}{VOC2012-FOG}
			& GradCAM
			& 0.1536 & 0.5626 & 0.1584 & 0.4292 & 0.4608 & 0.3305 & 1.5951 & \underline{1149.2333} & \underline{8705} \\
			& GradCAM++
			& 0.2024 & 0.5074 & 0.1924 & 0.4232 & 0.4805 & 0.3232 & 1.7016 & 1346.1962 & \underline{8705} \\
			& ScoreCAM
			& 0.2005 & 0.5370 & 0.1748 & 0.3969 & 0.6619 & \underline{0.6462} & \underline{0.6399} & 10375.5946 & \textbf{839} \\
			& DRISE
			& 0.2667 & 0.4225 & 0.2590 & \underline{0.3922} & \textbf{0.7549} & 0.2954 & 1.6886 & 48508.0483 & \textbf{839} \\
			& ODAM
			& \underline{0.2995} & \textbf{0.3838} & \textbf{0.2680} & \textbf{0.3651} & 0.6642 & 0.5463 & 0.8110 & 1474.6802 & \underline{8705} \\
			& PIEDet (ours)
			& \textbf{0.3074} & \underline{0.4059} & \underline{0.2616} & 0.3995 & \underline{0.6871} & \textbf{0.6642} & \textbf{0.5613} & \textbf{736.3962} & \textbf{839} \\
			
			\hline
		\end{tabular}
	}
\end{table*}

\subsection{Explainability Analysis}
\subsubsection{Quantitative Analysis}
To further validate the effectiveness of the explanation maps generated by PIEDet, we compare it with five mainstream post-hoc interpretability methods: GradCAM, GradCAM++, ScoreCAM, DRISE, and ODAM. Unlike these methods, PIEDet does not require additional backpropagation or extensive perturbation sampling after detection, but directly generates explanation maps from class-prototype responses along the forward inference path.

The quantitative results are reported in Table~\ref{tab:explainability_quantitative}. On ExDark, ODAM achieves the best results for Ins. Class, Del. Class, Ins. IoU, and Del. IoU. PIEDet ranks second or remains close to the best on these metrics while providing better performance in Energy PG, Comp., Time, and Memory. This indicates that PIEDet reduces explanation-generation cost while retaining high explanation quality. In particular, its runtime is lower than that of every post-hoc method, demonstrating the inference-efficiency advantage of intrinsic explanations.

On RTTS, PIEDet achieves the best Ins. Class, Ins. IoU, and Energy PG scores, showing that its explanation maps more effectively preserve regions important for category discrimination and localization. PIEDet also has the lowest runtime, at only 707.9164, substantially outperforming all competing methods. These results indicate that its prototype responses focus well on object regions and can be generated at much lower cost.

On VOC2012-FOG, PIEDet achieves the best results for Ins. Class, Energy PG, Comp., Time, and Memory, while remaining competitive in Del. Class, Ins. IoU, and Point Game. Compared with methods such as DRISE and ODAM, its explanations are more compact and require substantially less generation time. Although DRISE obtains a high Point Game score, its extensive random perturbation sampling makes its runtime far greater than that of other approaches. ScoreCAM provides strong compactness but similarly incurs high cost because of repeated forward passes. Overall, PIEDet achieves a superior balance among explanation quality, computational efficiency, and memory usage, validating the effectiveness of its intrinsic prototype-based explanation mechanism.

\subsubsection{Qualitative Analysis}
In addition to quantitative evaluation, we visualize explanation maps in low-light and hazy scenes to intuitively analyze PIEDet's intrinsic explanation capability. The visual comparisons include PIEDet, GradCAM, GradCAM++, ScoreCAM, DRISE, and ODAM.

\begin{figure*}[!t]
	\centering
	\includegraphics[width=\textwidth]{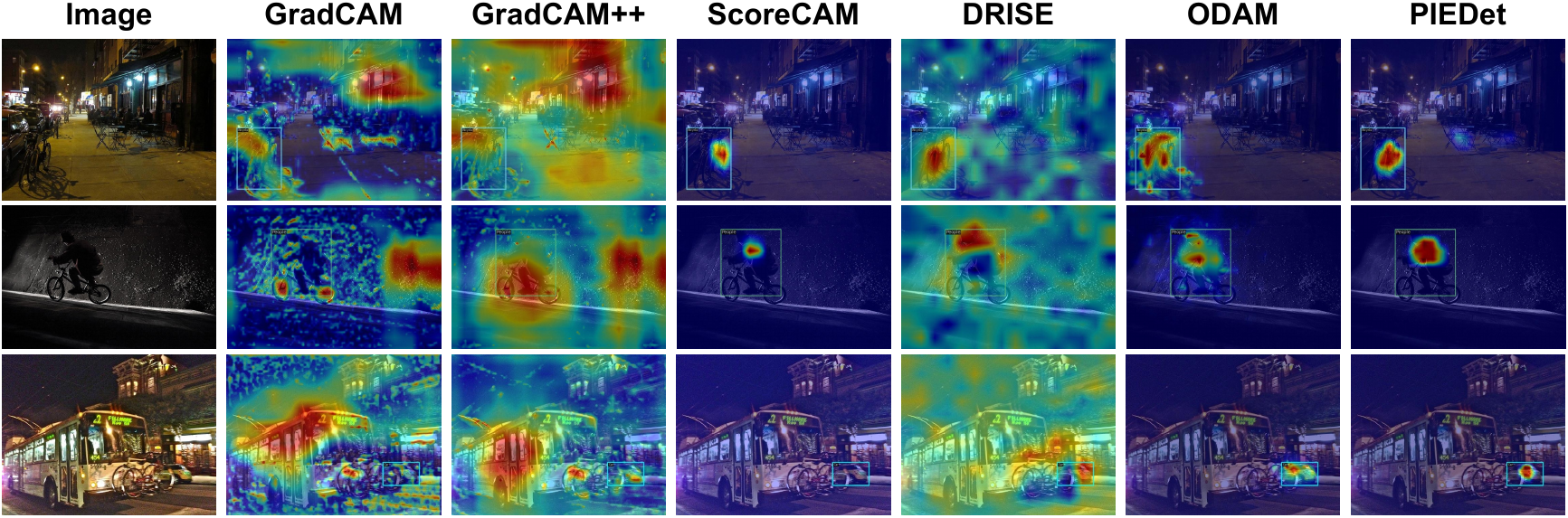}
	\caption{Visual comparison of explanation maps generated by PIEDet and mainstream post hoc interpretability methods in low-light scenes.}
	\label{fig:explanation-maps-low-light-scenes}
\end{figure*}

Figure~\ref{fig:explanation-maps-low-light-scenes} presents explanation maps in low-light scenes. The responses of GradCAM and GradCAM++ are generally diffuse and often cover large background areas around the object, indicating that their gradient attribution is susceptible to weak textures and noise. Although DRISE highlights parts of the object, its responses are often scattered and prone to spurious background activation.
ScoreCAM responses generally focus more strongly on object regions inside the detection box and exhibit less background diffusion. However, its responses to small objects are unstable, and object activation is missing in some samples. ODAM suppresses background interference relatively well, but its high-response regions within objects remain somewhat fragmented and fail to continuously cover the main object body.
In contrast, the explanation maps generated by PIEDet focus more consistently on the main object regions inside the detection boxes and effectively suppress background responses. This confirms that the proposed region-to-prototype matching loss establishes a clearer correspondence between class prototypes and ground-truth object regions, thereby improving object-region focus and explanation consistency.

\begin{figure*}[!t]
	\centering
	\includegraphics[width=\textwidth]{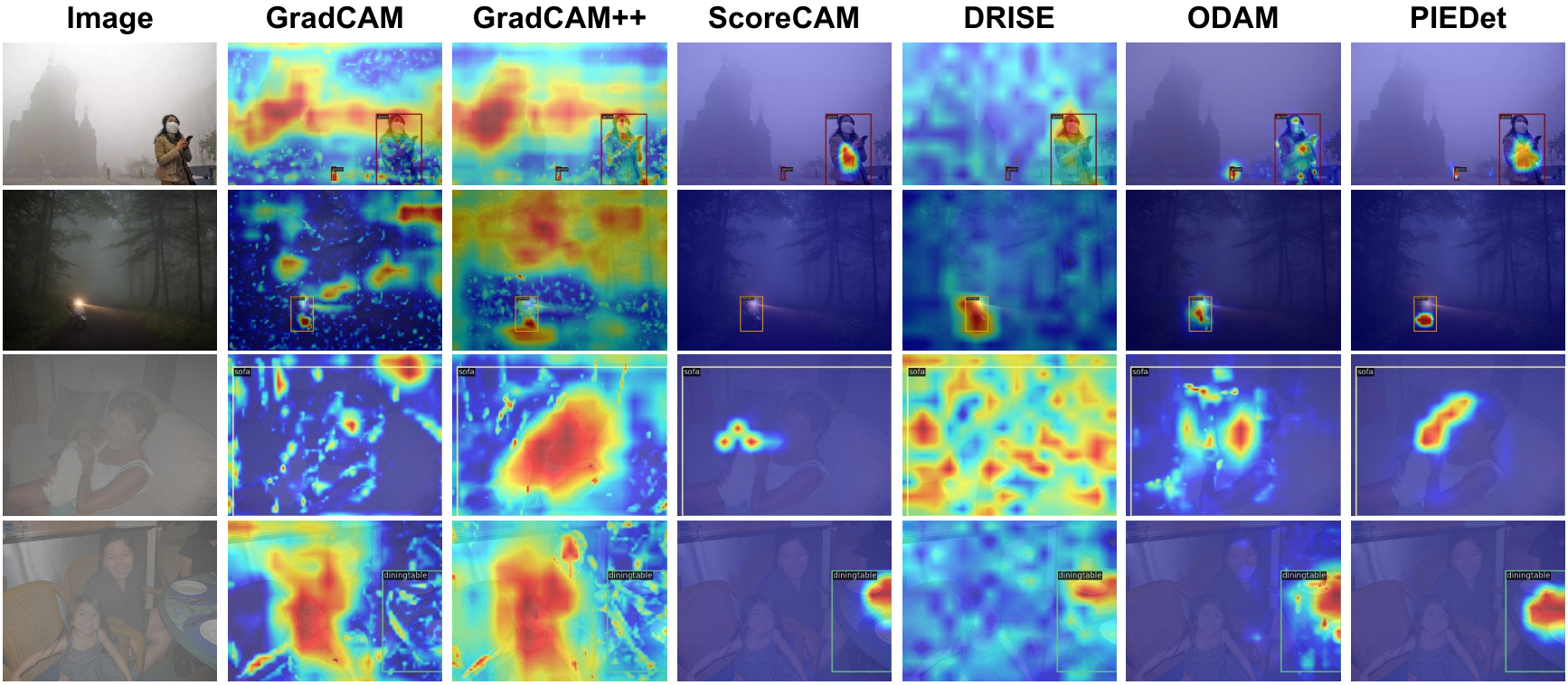}
	\caption{Visual comparison of explanation maps generated by PIEDet and mainstream post hoc interpretability methods in hazy scenes.}
	\label{fig:explanation-maps-hazy-scenes}
\end{figure*}

Figure~\ref{fig:explanation-maps-hazy-scenes} presents explanation maps in hazy scenes. GradCAM and GradCAM++ again produce diffuse activations that cover broad background regions around the object in some samples, making it difficult to obtain clear and stable discriminative evidence. DRISE can highlight the approximate object location, but its explanations are strongly affected by random perturbation sampling, resulting in granular and scattered responses as well as spurious background activation.
ScoreCAM produces more localized responses that generally converge toward object regions inside the detection boxes. Nevertheless, it can lose responses when explaining small objects, as illustrated by the second example in Fig.~\ref{fig:explanation-maps-hazy-scenes}. ODAM suppresses background interference relatively effectively, but its activations within the object remain fragmented, and the high-response regions do not continuously cover the principal object body.
Compared with these methods, PIEDet generates more compact and stable explanation maps, concentrating the dominant responses on the main object body or key discriminative regions inside the detection box while reducing irrelevant activations caused by hazy backgrounds. It therefore exhibits better spatial consistency and explanation stability.

Overall, PIEDet demonstrates strong explanation performance in both quantitative and qualitative evaluations. Its advantage over mainstream post hoc methods lies not only in producing explanation maps focused on object regions, but also in deriving those explanations directly from the detector's forward classification path without additional backpropagation or large-scale perturbation sampling. PIEDet can therefore generate detection results and discriminative evidence simultaneously with low memory and computational overhead.

\begin{table}[!t]
	\caption{Ablation results (mAP) of different module combinations on ExDark, RTTS, and VOC2012-FOG datasets.}
	\label{tab:ablation-modules}
	\centering
	\renewcommand{\arraystretch}{1.05}
	\resizebox{\columnwidth}{!}{
		\begin{tabular}{
				>{\centering\arraybackslash}m{1.5cm}
				>{\centering\arraybackslash}m{1.7cm}
				>{\centering\arraybackslash}m{1.2cm} |
				>{\centering\arraybackslash}m{1.2cm}
				>{\centering\arraybackslash}m{1.2cm}
				>{\centering\arraybackslash}m{2.5cm}}
			\hline
			\multicolumn{3}{c|}{{Modules}} & \multicolumn{3}{c}{{Datasets}} \\
			\hline
			\makecell[c]{PR-Loss} & \makecell[c]{RPM-Loss} & \makecell[c]{SAHPS} & ExDark & RTTS & VOC2012-FOG \\
			\hline
			& & & 64.4 & 74.6 & 58.9 \\
			& \checkmark & & 58.6 & 74.0 & 59.5 \\
			\checkmark & & & \underline{66.5} & 73.9 & \underline{62.8} \\
			\checkmark & \checkmark & & 64.9 & 74.2 & 62.2 \\
			& \checkmark & \checkmark & 65.1 & \underline{75.4} & 60.3 \\
			\checkmark & \checkmark & \checkmark & \textbf{69.1} & \textbf{76.0} & \textbf{63.7} \\
			\hline
		\end{tabular}
	}
\end{table}

\subsection{Ablation Study}
To systematically analyze the contribution of each design in the proposed framework, we conduct ablation studies on ExDark, RTTS, and VOC2012-FOG.

\subsubsection{Effects of Different Modules}
\label{sec_AS-Modules}
Table~\ref{tab:ablation-modules} compares different combinations of PR-Loss, RPM-Loss, and SAHPS. Introducing PR-Loss alone improves performance by 2.1 and 3.9 percentage points on ExDark and VOC2012-FOG, respectively, demonstrating its effectiveness under low-light and hazy conditions. In contrast, the gain from RPM-Loss alone is inconsistent. Adding SAHPS on top of RPM-Loss improves performance on all three datasets, indicating that the two components are complementary. When all modules are jointly used, the model achieves the best mAP values of 69.1, 76.0, and 63.7 on ExDark, RTTS, and VOC2012-FOG, respectively, improving over the baseline by 4.7, 1.4, and 4.8 percentage points. These results show that the modules work synergistically to improve detection across different low-visibility conditions.

\subsubsection{Effects of Different Feature Layers}
We further evaluate the role of each hierarchy level by using prototypes from only one feature level. The results in Table~\ref{tab:single_feature_levels} reveal clear scale preferences: the $8 \times 8$ level is more suitable for small objects, the $16 \times 16$ level is more effective for medium-sized objects, and the $32 \times 32$ level performs better for large objects. This behavior demonstrates that hierarchical prototype modeling effectively captures scale-dependent semantic information and verifies the necessity of the multi-level prototype design.

\begin{table*}[!t]
	\caption{Ablation results using single feature layers on ExDark, RTTS, and VOC2012-FOG datasets to analyze scale sensitivity.}
	\label{tab:single_feature_levels}
	\centering
	\renewcommand{\arraystretch}{1.05}  
	\resizebox{\textwidth}{!}{
		\begin{tabular}{
				>{\centering\arraybackslash}m{2.0cm} 
				>{\centering\arraybackslash}m{2.5cm} | 
				>{\centering\arraybackslash}m{1.7cm} 
				>{\centering\arraybackslash}m{1.9cm} 
				>{\centering\arraybackslash}m{1.9cm} | 
				>{\centering\arraybackslash}m{1.9cm}
				>{\centering\arraybackslash}m{1.9cm}
				>{\centering\arraybackslash}m{1.7cm} | 
				>{\centering\arraybackslash}m{1.7cm}
				>{\centering\arraybackslash}m{1.9cm}
				>{\centering\arraybackslash}m{1.9cm}
			}
			\hline
			\multirow{2}{*}[-1.0ex]{{Feature Level}} & \multirow{2}{*}[-1.0ex]{{Method}}
			& \multicolumn{3}{c|}{{ExDark}}
			& \multicolumn{3}{c|}{{RTTS}}
			& \multicolumn{3}{c}{{VOC2012-FOG}} \\
			\cline{3-5} \cline{6-8} \cline{9-11}
			& & mAP\textsubscript{s} & mAP\textsubscript{m} & mAP\textsubscript{l} 
			& mAP\textsubscript{s} & mAP\textsubscript{m} & mAP\textsubscript{l} 
			& mAP\textsubscript{s} & mAP\textsubscript{m} & mAP\textsubscript{l} \\
			\hline
			\multirow{2}{*}{Layer1 ($8 \times 8$)} & YOLOv8 & {3.9} & 4.2 & 0.5 & {20.1} & 24.1 & 1.4 & {1.1} & 4.3 & 0.0 \\
			& PIEDet (ours) & \textbf{4.2 (↑0.3)} & 2.5 & 0.2 & \textbf{21.0 (↑0.9)} & 12.6 & 0.2 & \textbf{1.8 (↑0.7)} & 2.2 & 0.0 \\
			\hline
			\multirow{2}{*}{Layer2 ($16 \times 16$)} & YOLOv8 & 0.0 & {14.9} & 7.6 & 5.5 & {39.2} & \textbf{41.3} & 0.0 & {8.6} & 7.9 \\
			& PIEDet (ours) & 1.7 & \textbf{23.3 (↑8.4)} & 9.6 & 12.5 & \textbf{48.5 (↑9.3)} & 38.2 & 0.1 & \textbf{15.7 (↑7.1)} & 8.9 \\
			\hline
			\multirow{2}{*}{Layer3 ($32 \times 32$)} & YOLOv8 & 0.0 & 4.9 & {29.0} & 0.0 & 1.3 & 33.8 & 0.0 & 0.1 & {33.5} \\
			& PIEDet (ours) & 0.0 & 6.1 & \textbf{36.6 (↑7.6)} & 0.0 & 2.4 & {38.5 (↑4.7)} & 0.0 & 1.1 & \textbf{42.9 (↑9.4)} \\
			\hline
		\end{tabular}
	}
\end{table*}

\subsubsection{Effects of Hyperparameter Settings}
We analyze the effects of different hyperparameter configurations $(\tau_1,\tau_2,\tau_3)$ from the perspectives of computational efficiency and detection accuracy, focusing on how hierarchical prototype allocation affects computational cost, inference speed, and detection performance.

Table~\ref{tab:ablation-hyperparameters-efficiency} reports the efficiency results under different hyperparameter settings. As $(\tau_1,\tau_2)$ increase, the parameter count and computational cost gradually rise, leading to a reduction in FPS. Excessively large settings therefore reduce inference efficiency, indicating that appropriate hierarchical allocation is important for balancing computational overhead and representation capacity.

Table~\ref{tab:ablation-hyperparameters-performance} reports detection accuracy under different settings. The best performance is obtained when $\tau_1=4$, $\tau_2=8$, and $\tau_3=H/s_3$, indicating that an appropriate scale partition better matches the receptive fields of different feature levels. Here, $H$ is the input image height and $s_3$ is the downsampling stride of the final feature map.

\begin{table*}[!t]
	\caption{Ablation study on the hyperparameter settings ($\tau_1,\tau_2,\tau_3$) evaluating efficiency metrics on FPS, FLOPs, and Params. $H$ denotes the input image height and $s_3$ is the stride of the final feature map.}
	\label{tab:ablation-hyperparameters-efficiency}
	\centering
	\renewcommand{\arraystretch}{1.05}
	\resizebox{\textwidth}{!}{
		\begin{tabular}{
				>{\centering\arraybackslash}m{6.0cm} |
				>{\centering\arraybackslash}m{5.0cm}
				>{\centering\arraybackslash}m{5.0cm}
				>{\centering\arraybackslash}m{5.0cm}
			}
			\hline
			{Method} & FPS $\uparrow$ & FLOPs (G) $\downarrow$ & Params (M) $\downarrow$ \\
			\hline
			YOLOv8 & 132.67 & 22.317 & 11.164 \\
			Ours ($\tau_1{=}2$, $\tau_2{=}4$, $\tau_3{=}H/s_3$) & 138.92 \textcolor{blue}{(↑ \textbf{4.711}\%)} & 21.262 \textcolor{blue}{(↓ \textbf{4.727}\%)} & 10.815 \textcolor{blue}{(↓ \textbf{3.126}\%)} \\
			Ours ($\tau_1{=}4$, $\tau_2{=}8$, $\tau_3{=}H/s_3$) & 137.23 \textcolor{blue}{(↑ \textbf{3.437}\%)} & 21.266 \textcolor{blue}{(↓ \textbf{4.709}\%)} & 10.815 \textcolor{blue}{(↓ \textbf{3.126}\%)} \\
			Ours ($\tau_1{=}8$, $\tau_2{=}16$, $\tau_3{=}H/s_3$) & 135.73 \textcolor{blue}{(↑ \textbf{2.306}\%)} & 21.274 \textcolor{blue}{(↓ \textbf{4.673}\%)} & 10.817 \textcolor{blue}{(↓ \textbf{3.108}\%)} \\
			\hline
		\end{tabular}
	}
\end{table*}

\begin{table}[!t]
	\caption{Effects of hyperparameter settings across ExDark, RTTS, and VOC2012-FOG datasets. $H$ denotes the input image height and $s_3$ is the stride of the final feature map.}
	\label{tab:ablation-hyperparameters-performance}
	\centering
	\renewcommand{\arraystretch}{1.05}
	\resizebox{\columnwidth}{!}{
		\begin{tabular}{
				>{\centering\arraybackslash}m{1.0cm}
				>{\centering\arraybackslash}m{1.0cm}
				>{\centering\arraybackslash}m{1.0cm} |
				>{\centering\arraybackslash}m{1.0cm}
				>{\centering\arraybackslash}m{1.0cm}
				>{\centering\arraybackslash}m{2.5cm}}
			\hline
			\makecell[c]{$\tau_1$} & \makecell[c]{$\tau_2$} & \makecell[c]{$\tau_3$}
			& ExDark & RTTS & VOC2012-FOG \\
			\hline
			2 & 4 & $H/s_3$ & 66.4 & 74.5 & 58.8 \\
			4 & 8 & $H/s_3$ & \textbf{69.1} & \textbf{76.0} & \textbf{63.7} \\
			8 & 16 & $H/s_3$ & \underline{67.8} & \underline{75.2} & \underline{62.9} \\
			\hline
		\end{tabular}
	}
\end{table}

\begin{table}[!t]
	\caption{Comparison (mAP@0.5) of different prototype regularization methods on ExDark, RTTS, and VOC2012-FOG datasets.}
	\label{tab:pr_loss_comparison}
	\centering
	\renewcommand{\arraystretch}{1.05}
	\setlength{\tabcolsep}{10pt}
	\resizebox{\columnwidth}{!}{
		\begin{tabular}{
				>{\centering\arraybackslash}m{2.5cm} |
				>{\centering\arraybackslash}m{1.5cm}
				>{\centering\arraybackslash}m{1.5cm}
				>{\centering\arraybackslash}m{2.5cm}
			}
			\hline
			{Methods} & ExDark & RTTS & VOC2012-FOG \\
			\hline
			Cosine & 62.9 & 75.4 & 54.7 \\
			POP~\cite{ref_POP} & \underline{63.7} & \underline{75.6} & \underline{59.4} \\
			SVD (Ours) & \textbf{69.1} & \textbf{76.0} & \textbf{63.7} \\
			\hline
		\end{tabular}
	}
\end{table}

\subsubsection{Effectiveness of Prototype Regularization Loss}
To evaluate the influence of different constraint strategies on prototype regularization, we compare three implementations: a cosine-based loss, POP, and our SVD-based global spectral constraint. The quantitative results are shown in Table~\ref{tab:pr_loss_comparison}. The SVD constraint achieves the highest mAP on all three datasets, indicating that constraining prototype independence from the perspective of the entire prototype space is more effective than relying solely on pairwise constraints.

Among the two pairwise constraints, POP outperforms the cosine-based constraint, suggesting that a soft inner-product penalty provides more stable regularization. Nevertheless, compared with our SVD-based global constraint, the performance gains of pairwise approaches remain smaller and less consistent. Overall, the SVD-based prototype regularization loss more effectively optimizes the structure of the prototype space and improves detection performance under diverse challenging imaging conditions.

\section{Conclusion}
In this paper, we propose PIEDet, a prototype-driven intrinsically explainable object detection framework that integrates interpretable prototype learning into the detection process, allowing the model to generate intrinsic discriminative evidence together with detection results. Unlike interpretability methods based on post-hoc attribution, PIEDet formulates category discrimination as matching between region features and class prototypes and constructs hierarchical class prototypes at different detection levels of a multi-scale feature pyramid, thereby learning scale-aware class-semantic representations. To improve the discriminability and regional focus of prototype representations, we further introduce a prototype-driven feature optimization method comprising a prototype regularization loss and a region-to-prototype matching loss. These losses reduce redundancy among class prototypes and align object-region features with class prototypes, respectively. We also propose a scale-aligned hierarchical prototype supervision mechanism that assigns scale-matched regional supervision signals to different detection levels, reducing training noise caused by cross-scale objects. Experiments on ExDark, RTTS, and VOC2012-FOG demonstrate that PIEDet improves low-quality object detection without relying on image enhancement or complex architectural design, while generating intrinsic explanatory evidence focused on object regions. Compared with mainstream post-hoc interpretability methods, PIEDet provides a better balance among explanation quality, computational cost, and memory consumption, validating the effectiveness of the proposed prototype-driven intrinsically explainable detection framework.

\bibliographystyle{IEEEtran}
\bibliography{PIEDet}

@inproceedings{ref_AD1,
	title={Deepdriving: Learning affordance for direct perception in autonomous driving},
	author={Chen, Chenyi and Seff, Ari and Kornhauser, Alain and Xiao, Jianxiong},
	booktitle={Proceedings of the IEEE international conference on computer vision},
	pages={2722--2730},
	year={2015}
}

@article{ref_AD2,
	title={Multi-attention DenseNet: A scattering medium imaging optimization framework for visual data pre-processing of autonomous driving systems},
	author={Liu, Peng and Zhang, Chufeng and Qi, Hao and Wang, Guoyu and Zheng, Haiyong},
	journal={IEEE Transactions on Intelligent Transportation Systems},
	volume={23},
	number={12},
	pages={25396--25407},
	year={2022},
	publisher={IEEE}
}

@inproceedings{ref_CAM,
	title={Learning deep features for discriminative localization},
	author={Zhou, Bolei and Khosla, Aditya and Lapedriza, Agata and Oliva, Aude and Torralba, Antonio},
	booktitle={Proceedings of the IEEE conference on computer vision and pattern recognition},
	pages={2921--2929},
	year={2016}
}

@inproceedings{ref_GradCAM,
	title={Grad-cam: Visual explanations from deep networks via gradient-based localization},
	author={Selvaraju, Ramprasaath R and Cogswell, Michael and Das, Abhishek and Vedantam, Ramakrishna and Parikh, Devi and Batra, Dhruv},
	booktitle={Proceedings of the IEEE international conference on computer vision},
	pages={618--626},
	year={2017}
}

@inproceedings{ref_GradCAMpp,
	title={Grad-cam++: Generalized gradient-based visual explanations for deep convolutional networks},
	author={Chattopadhay, Aditya and Sarkar, Anirban and Howlader, Prantik and Balasubramanian, Vineeth N},
	booktitle={2018 IEEE winter conference on applications of computer vision (WACV)},
	pages={839--847},
	year={2018},
	organization={IEEE}
}

@article{ref_LayerCAM,
	title={Layercam: Exploring hierarchical class activation maps for localization},
	author={Jiang, Peng-Tao and Zhang, Chang-Bin and Hou, Qibin and Cheng, Ming-Ming and Wei, Yunchao},
	journal={IEEE transactions on image processing},
	volume={30},
	pages={5875--5888},
	year={2021},
	publisher={IEEE}
}

@inproceedings{ref_FinerCAM,
	title={Finer-cam: Spotting the difference reveals finer details for visual explanation},
	author={Zhang, Ziheng and Gu, Jianyang and Chowdhury, Arpita and Mai, Zheda and Carlyn, David and Berger-Wolf, Tanya and Su, Yu and Chao, Wei-Lun},
	booktitle={Proceedings of the Computer Vision and Pattern Recognition Conference},
	pages={9611--9620},
	year={2025}
}

@inproceedings{ref_ScoreCAM,
	title={Score-CAM: Score-weighted visual explanations for convolutional neural networks},
	author={Wang, Haofan and Wang, Zifan and Du, Mengnan and Yang, Fan and Zhang, Zijian and Ding, Sirui and Mardziel, Piotr and Hu, Xia},
	booktitle={Proceedings of the IEEE/CVF conference on computer vision and pattern recognition workshops},
	pages={24--25},
	year={2020}
}

@inproceedings{ref_SSGradCAMpp,
	title={Spatial Sensitive Grad-CAM++: Improved visual explanation for object detectors via weighted combination of gradient map},
	author={Yamauchi, Toshinori},
	booktitle={Proceedings of the IEEE/CVF Conference on Computer Vision and Pattern Recognition},
	pages={8164--8168},
	year={2024}
}

@inproceedings{ref_DRISE,
	title={Black-box explanation of object detectors via saliency maps},
	author={Petsiuk, Vitali and Jain, Rajiv and Manjunatha, Varun and Morariu, Vlad I and Mehra, Ashutosh and Ordonez, Vicente and Saenko, Kate},
	booktitle={Proceedings of the IEEE/CVF conference on computer vision and pattern recognition},
	pages={11443--11452},
	year={2021}
}

@article{ref_ODAM,
	title={Gradient-based instance-specific visual explanations for object specification and object discrimination},
	author={Zhao, Chenyang and Hsiao, Janet H and Chan, Antoni B},
	journal={IEEE Transactions on Pattern Analysis and Machine Intelligence},
	volume={46},
	number={9},
	pages={5967--5985},
	year={2024},
	publisher={IEEE}
}

@article{ref_ProtoPNet,
	title={This looks like that: deep learning for interpretable image recognition},
	author={Chen, Chaofan and Li, Oscar and Tao, Daniel and Barnett, Alina and Rudin, Cynthia and Su, Jonathan K},
	journal={Advances in neural information processing systems},
	volume={32},
	year={2019}
}

@inproceedings{ref_segmentation1,
	title={CSC-PA: Cross-image Semantic Correlation via Prototype Attentions for Single-network Semi-supervised Breast Tumor Segmentation},
	author={Ding, Zhenhui and Chen, Guilian and Zhang, Qin and Wu, Huisi and Qin, Jing},
	booktitle={Proceedings of the Computer Vision and Pattern Recognition Conference},
	pages={15632--15641},
	year={2025}
}

@inproceedings{ref_segmentation2,
	title={Prototype-based image prompting for weakly supervised histopathological image segmentation},
	author={Tang, Qingchen and Fan, Lei and Pagnucco, Maurice and Song, Yang},
	booktitle={Proceedings of the Computer Vision and Pattern Recognition Conference},
	pages={30271--30280},
	year={2025}
}

@inproceedings{ref_segmentation3,
	title={Weakly supervised semantic segmentation by pixel-to-prototype contrast},
	author={Du, Ye and Fu, Zehua and Liu, Qingjie and Wang, Yunhong},
	booktitle={Proceedings of the IEEE/CVF conference on computer vision and pattern recognition},
	pages={4320--4329},
	year={2022}
}

@inproceedings{ref_FSOD1,
	title={Universal-prototype enhancing for few-shot object detection},
	author={Wu, Aming and Han, Yahong and Zhu, Linchao and Yang, Yi},
	booktitle={Proceedings of the IEEE/CVF international conference on computer vision},
	pages={9567--9576},
	year={2021}
}

@inproceedings{ref_FSOD2,
	title={Few-shot object detection by attending to per-sample-prototype},
	author={Lee, Hojun and Lee, Myunggi and Kwak, Nojun},
	booktitle={Proceedings of the IEEE/CVF Winter Conference on Applications of Computer Vision},
	pages={2445--2454},
	year={2022}
}

@article{ref_FSOD3,
	title={Few-shot cross-domain object detection with instance-level prototype-based meta-learning},
	author={Zhang, Lin and Zhang, Bo and Shi, Botian and Fan, Jiayuan and Chen, Tao},
	journal={IEEE Transactions on Circuits and Systems for Video Technology},
	volume={34},
	number={10},
	pages={9078--9089},
	year={2024},
	publisher={IEEE}
}

@article{ref_FSOD4,
	title={A Generalized Few-Shot Object Detection Method via Extraction of Base-Novel Commonality with Memory Distillation of Category Prototypes},
	author={Su, Junchi and Gao, Xin and Lu, Heping and Li, Baofeng and Zhai, Feng and Fang, Xiao and Wang, Taizhi and Li, Qiangwei},
	journal={IEEE Transactions on Circuits and Systems for Video Technology},
	year={2025},
	publisher={IEEE}
}

@inproceedings{ref_RCNN,
	title={Rich feature hierarchies for accurate object detection and semantic segmentation},
	author={Girshick, Ross and Donahue, Jeff and Darrell, Trevor and Malik, Jitendra},
	booktitle={Proceedings of the IEEE conference on computer vision and pattern recognition},
	pages={580--587},
	year={2014}
}

@article{ref_SPPNet,
	title={Spatial pyramid pooling in deep convolutional networks for visual recognition},
	author={He, Kaiming and Zhang, Xiangyu and Ren, Shaoqing and Sun, Jian},
	journal={IEEE transactions on pattern analysis and machine intelligence},
	volume={37},
	number={9},
	pages={1904--1916},
	year={2015},
	publisher={IEEE}
}

@inproceedings{ref_FastRCNN,
	title={Fast r-cnn},
	author={Girshick, Ross},
	booktitle={Proceedings of the IEEE international conference on computer vision},
	pages={1440--1448},
	year={2015}
}

@article{ref_FasterRCNN,
	title={Faster R-CNN: Towards real-time object detection with region proposal networks},
	author={Ren, Shaoqing and He, Kaiming and Girshick, Ross and Sun, Jian},
	journal={IEEE transactions on pattern analysis and machine intelligence},
	volume={39},
	number={6},
	pages={1137--1149},
	year={2016},
	publisher={IEEE}
}

@inproceedings{ref_YOLO,
	title={You only look once: Unified, real-time object detection},
	author={Redmon, Joseph and Divvala, Santosh and Girshick, Ross and Farhadi, Ali},
	booktitle={Proceedings of the IEEE conference on computer vision and pattern recognition},
	pages={779--788},
	year={2016}
}

@inproceedings{ref_CornerNet,
	title={Cornernet: Detecting objects as paired keypoints},
	author={Law, Hei and Deng, Jia},
	booktitle={Proceedings of the European conference on computer vision (ECCV)},
	pages={734--750},
	year={2018}
}

@inproceedings{ref_ExtremeNet,
	title={Bottom-up object detection by grouping extreme and center points},
	author={Zhou, Xingyi and Zhuo, Jiacheng and Krahenbuhl, Philipp},
	booktitle={Proceedings of the IEEE/CVF conference on computer vision and pattern recognition},
	pages={850--859},
	year={2019}
}

@inproceedings{ref_CenterNet,
	title={Centernet: Keypoint triplets for object detection},
	author={Duan, Kaiwen and Bai, Song and Xie, Lingxi and Qi, Honggang and Huang, Qingming and Tian, Qi},
	booktitle={Proceedings of the IEEE/CVF international conference on computer vision},
	pages={6569--6578},
	year={2019}
}

@inproceedings{ref_DETR,
	title={End-to-end object detection with transformers},
	author={Carion, Nicolas and Massa, Francisco and Synnaeve, Gabriel and Usunier, Nicolas and Kirillov, Alexander and Zagoruyko, Sergey},
	booktitle={European conference on computer vision},
	pages={213--229},
	year={2020},
	organization={Springer}
}

@inproceedings{ref_SAMDETR,
	title={Accelerating DETR convergence via semantic-aligned matching},
	author={Zhang, Gongjie and Luo, Zhipeng and Yu, Yingchen and Cui, Kaiwen and Lu, Shijian},
	booktitle={Proceedings of the IEEE/CVF conference on computer vision and pattern recognition},
	pages={949--958},
	year={2022}
}

@article{ref_DeformableDETR,
title={Deformable detr: Deformable transformers for end-to-end object detection},
author={Zhu, Xizhou and Su, Weijie and Lu, Lewei and Li, Bin and Wang, Xiaogang and Dai, Jifeng},
journal={arXiv preprint arXiv:2010.04159},
year={2020}
}

@inproceedings{ref_LLIE1,
	title={Aglldiff: Guiding diffusion models towards unsupervised training-free real-world low-light image enhancement},
	author={Lin, Yunlong and Ye, Tian and Chen, Sixiang and Fu, Zhenqi and Wang, Yingying and Chai, Wenhao and Xing, Zhaohu and Li, Wenxue and Zhu, Lei and Ding, Xinghao},
	booktitle={Proceedings of the AAAI Conference on Artificial Intelligence},
	volume={39},
	number={5},
	pages={5307--5315},
	year={2025}
}

@inproceedings{ref_LLIE2,
	title={Hvi: A new color space for low-light image enhancement},
	author={Yan, Qingsen and Feng, Yixu and Zhang, Cheng and Pang, Guansong and Shi, Kangbiao and Wu, Peng and Dong, Wei and Sun, Jinqiu and Zhang, Yanning},
	booktitle={Proceedings of the Computer Vision and Pattern Recognition Conference},
	pages={5678--5687},
	year={2025}
}

@inproceedings{ref_DEHAZING1,
	title={Beyond spatial domain: Cross-domain promoted fourier convolution helps single image dehazing},
	author={Zhang, Xiaozhe and Ding, Haidong and Xie, Fengying and Pan, Linpeng and Zi, Yue and Wang, Ke and Zhang, Haopeng},
	booktitle={Proceedings of the AAAI Conference on Artificial Intelligence},
	volume={39},
	number={10},
	pages={10221--10229},
	year={2025}
}

@inproceedings{ref_DEHAZING2,
	title={Exploiting diffusion prior for real-world image dehazing with unpaired training},
	author={Lan, Yunwei and Cui, Zhigao and Liu, Chang and Peng, Jialun and Wang, Nian and Luo, Xin and Liu, Dong},
	booktitle={Proceedings of the AAAI Conference on Artificial Intelligence},
	volume={39},
	number={4},
	pages={4455--4463},
	year={2025}
}

@inproceedings{ref_DEHAZING3,
	title={Prior-guided hierarchical harmonization network for efficient image dehazing},
	author={Su, Xiongfei and Li, Siyuan and Cui, Yuning and Cao, Miao and Zhang, Yulun and Chen, Zheng and Wu, Zongliang and Wang, Zedong and Zhang, Yuanlong and Yuan, Xin},
	booktitle={Proceedings of the AAAI Conference on Artificial Intelligence},
	volume={39},
	number={7},
	pages={7042--7050},
	year={2025}
}

@inproceedings{ref_DEHAZING4,
	title={Dehaze-RetinexGAN: Real-World Image Dehazing via Retinex-based Generative Adversarial Network},
	author={Wang, Xinran and Yang, Guang and Ye, Tian and Liu, Yun},
	booktitle={Proceedings of the AAAI Conference on Artificial Intelligence},
	volume={39},
	number={8},
	pages={7997--8005},
	year={2025}
}

@inproceedings{ref_HLA-Face,
	title={Hla-face: Joint high-low adaptation for low light face detection},
	author={Wang, Wenjing and Yang, Wenhan and Liu, Jiaying},
	booktitle={Proceedings of the IEEE/CVF Conference on Computer Vision and Pattern Recognition},
	pages={16195--16204},
	year={2021}
}

@inproceedings{ref_MAET,
	title={Multitask aet with orthogonal tangent regularity for dark object detection},
	author={Cui, Ziteng and Qi, Guo-Jun and Gu, Lin and You, Shaodi and Zhang, Zenghui and Harada, Tatsuya},
	booktitle={Proceedings of the IEEE/CVF international conference on computer vision},
	pages={2553--2562},
	year={2021}
}

@article{ref_DSNet,
	title={DSNet: Joint semantic learning for object detection in inclement weather conditions},
	author={Huang, Shih-Chia and Le, Trung-Hieu and Jaw, Da-Wei},
	journal={IEEE transactions on pattern analysis and machine intelligence},
	volume={43},
	number={8},
	pages={2623--2633},
	year={2020},
	publisher={IEEE}
}

@inproceedings{ref_IA-YOLO,
	title={Image-adaptive YOLO for object detection in adverse weather conditions},
	author={Liu, Wenyu and Ren, Gaofeng and Yu, Runsheng and Guo, Shi and Zhu, Jianke and Zhang, Lei},
	booktitle={Proceedings of the AAAI conference on artificial intelligence},
	volume={36},
	number={2},
	pages={1792--1800},
	year={2022}
}

@inproceedings{ref_FeatEnHancer,
	title={Featenhancer: Enhancing hierarchical features for object detection and beyond under low-light vision},
	author={Hashmi, Khurram Azeem and Kallempudi, Goutham and Stricker, Didier and Afzal, Muhammad Zeshan},
	booktitle={Proceedings of the IEEE/CVF International Conference on Computer Vision},
	pages={6725--6735},
	year={2023}
}

@inproceedings{ref_T2,
	title={Trash to treasure: Low-light object detection via decomposition-and-aggregation},
	author={Cui, Xiaohan and Ma, Long and Ma, Tengyu and Liu, Jinyuan and Fan, Xin and Liu, Risheng},
	booktitle={Proceedings of the AAAI Conference on Artificial Intelligence},
	volume={38},
	number={2},
	pages={1417--1425},
	year={2024}
}

@article{ref_YOLA,
	title={You Only Look Around: Learning Illumination-Invariant Feature for Low-light Object Detection},
	author={Hong, Mingbo and Cheng, Shen and Huang, Haibin and Fan, Haoqiang and Liu, Shuaicheng},
	journal={Advances in Neural Information Processing Systems},
	volume={37},
	pages={87136--87158},
	year={2024}
}

@inproceedings{ref_MBLLEN,
	title={MBLLEN: Low-light image/video enhancement using cnns.},
	author={Lv, Feifan and Lu, Feng and Wu, Jianhua and Lim, Chongsoon},
	booktitle={Bmvc},
	volume={220},
	number={1},
	pages={4},
	year={2018},
	organization={Northumbria University}
}

@inproceedings{ref_KinD,
	title={Kindling the darkness: A practical low-light image enhancer},
	author={Zhang, Yonghua and Zhang, Jiawan and Guo, Xiaojie},
	booktitle={Proceedings of the 27th ACM international conference on multimedia},
	pages={1632--1640},
	year={2019}
}

@inproceedings{ref_Zero-DCE,
	title={Zero-reference deep curve estimation for low-light image enhancement},
	author={Guo, Chunle and Li, Chongyi and Guo, Jichang and Loy, Chen Change and Hou, Junhui and Kwong, Sam and Cong, Runmin},
	booktitle={Proceedings of the IEEE/CVF conference on computer vision and pattern recognition},
	pages={1780--1789},
	year={2020}
}

@inproceedings{ref_SCI,
	title={Toward fast, flexible, and robust low-light image enhancement},
	author={Ma, Long and Ma, Tengyu and Liu, Risheng and Fan, Xin and Luo, Zhongxuan},
	booktitle={Proceedings of the IEEE/CVF conference on computer vision and pattern recognition},
	pages={5637--5646},
	year={2022}
}

@inproceedings{ref_Retinexformer,
	title={Retinexformer: One-stage retinex-based transformer for low-light image enhancement},
	author={Cai, Yuanhao and Bian, Hao and Lin, Jing and Wang, Haoqian and Timofte, Radu and Zhang, Yulun},
	booktitle={Proceedings of the IEEE/CVF international conference on computer vision},
	pages={12504--12513},
	year={2023}
}

@inproceedings{ref_DiffUIR,
	title={Selective hourglass mapping for universal image restoration based on diffusion model},
	author={Zheng, Dian and Wu, Xiao-Ming and Yang, Shuzhou and Zhang, Jian and Hu, Jian-Fang and Zheng, Wei-Shi},
	booktitle={Proceedings of the IEEE/CVF Conference on Computer Vision and Pattern Recognition},
	pages={25445--25455},
	year={2024}
}

@article{ref_Diff-Retinex++,
	title={Diff-Retinex++: Retinex-Driven Reinforced Diffusion Model for Low-Light Image Enhancement},
	author={Yi, Xunpeng and Xu, Han and Zhang, Hao and Tang, Linfeng and Ma, Jiayi},
	journal={IEEE Transactions on Pattern Analysis and Machine Intelligence},
	year={2025},
	publisher={IEEE}
}

@inproceedings{ref_DiffDehaze,
	title={Learning Hazing to Dehazing: Towards Realistic Haze Generation for Real-World Image Dehazing},
	author={Wang, Ruiyi and Zheng, Yushuo and Zhang, Zicheng and Li, Chunyi and Liu, Shuaicheng and Zhai, Guangtao and Liu, Xiaohong},
	booktitle={Proceedings of the Computer Vision and Pattern Recognition Conference},
	pages={23091--23100},
	year={2025}
}

@article{ref_DNMGDT,
	title={Real scene single image dehazing network with multi-prior guidance and domain transfer},
	author={Su, Yanzhao and Wang, Nian and Cui, Zhigao and Cai, Yanping and He, Chuan and Li, Aihua},
	journal={IEEE Transactions on Multimedia},
	year={2025},
	publisher={IEEE}
}

@inproceedings{ref_SGDN,
	title={Guided real image dehazing using ycbcr color space},
	author={Fang, Wenxuan and Fan, Junkai and Zheng, Yu and Weng, Jiangwei and Tai, Ying and Li, Jun},
	booktitle={Proceedings of the AAAI Conference on Artificial Intelligence},
	volume={39},
	number={3},
	pages={2906--2914},
	year={2025}
}

@article{ref_MASFNet,
	title={MASFNet: Multi-scale Adaptive Sampling Fusion Network for Object Detection in Adverse Weather},
	author={Liu, Zhenbing and Fang, Tianle and Lu, Haoxiang and Zhang, Weidong and Lan, Rushi},
	journal={IEEE Transactions on Geoscience and Remote Sensing},
	year={2025},
	publisher={IEEE}
}

@article{ref_DFL,
	title={Generalized focal loss: Learning qualified and distributed bounding boxes for dense object detection},
	author={Li, Xiang and Wang, Wenhai and Wu, Lijun and Chen, Shuo and Hu, Xiaolin and Li, Jun and Tang, Jinhui and Yang, Jian},
	journal={Advances in neural information processing systems},
	volume={33},
	pages={21002--21012},
	year={2020}
}

@inproceedings{ref_OWN,
	title={Orthogonal weight normalization: Solution to optimization over multiple dependent stiefel manifolds in deep neural networks},
	author={Huang, Lei and Liu, Xianglong and Lang, Bo and Yu, Adams and Wang, Yongliang and Li, Bo},
	booktitle={Proceedings of the AAAI Conference on Artificial Intelligence},
	volume={32},
	number={1},
	year={2018}
}

@inproceedings{ref_POP,
title={Learning orthogonal prototypes for generalized few-shot semantic segmentation},
author={Liu, Sun-Ao and Zhang, Yiheng and Qiu, Zhaofan and Xie, Hongtao and Zhang, Yongdong and Yao, Ting},
booktitle={Proceedings of the IEEE/CVF conference on computer vision and pattern recognition},
pages={11319--11328},
year={2023}
}

@article{ref_TDN,
	title={Training deep networks with structured layers by matrix backpropagation},
	author={Ionescu, Catalin and Vantzos, Orestis and Sminchisescu, Cristian},
	journal={arXiv preprint arXiv:1509.07838},
	year={2015}
}

@article{ref_exdark,
	title={Getting to know low-light images with the exclusively dark dataset},
	author={Loh, Yuen Peng and Chan, Chee Seng},
	journal={Computer vision and image understanding},
	volume={178},
	pages={30--42},
	year={2019},
	publisher={Elsevier}
}

@article{ref_rtts,
	title={Benchmarking single-image dehazing and beyond},
	author={Li, Boyi and Ren, Wenqi and Fu, Dengpan and Tao, Dacheng and Feng, Dan and Zeng, Wenjun and Wang, Zhangyang},
	journal={IEEE transactions on image processing},
	volume={28},
	number={1},
	pages={492--505},
	year={2018},
	publisher={IEEE}
}

@article{ref_voc,
	title={The pascal visual object classes (voc) challenge},
	author={Everingham, Mark and Van Gool, Luc and Williams, Christopher KI and Winn, John and Zisserman, Andrew},
	journal={International journal of computer vision},
	volume={88},
	number={2},
	pages={303--338},
	year={2010},
	publisher={Springer}
}

\end{document}